\documentclass[11pt]{article}
\usepackage[preprint]{acl}

\usepackage{times}
\usepackage{latexsym}

\usepackage[T1]{fontenc}

\usepackage[utf8]{inputenc}

\usepackage{microtype}

\usepackage{inconsolata}

\usepackage{graphicx}
\usepackage{xcolor}
\usepackage{colortbl}
\usepackage{amsmath}
\usepackage{fvextra}
\usepackage{mdframed}
\usepackage{booktabs}
\usepackage{arydshln}
\usepackage{subcaption}

\newcommand{\prompttitle}[1]{\vspace{0.6em}\noindent{\footnotesize #1}\par\vspace{0.15em}}

\definecolor{promptbg}{HTML}{F5F7FA}

\newmdenv[%
  backgroundcolor=promptbg,%
  hidealllines=true,%
  innertopmargin=6pt,%
  innerbottommargin=6pt,%
  innerleftmargin=6pt,%
  innerrightmargin=6pt,%
  skipabove=2pt,%
  skipbelow=2pt%
]{examplebox}

\newenvironment{promptblock}
  {%
    \VerbatimEnvironment
    \begin{mdframed}[%
      backgroundcolor=promptbg,%
      hidealllines=true,%
      innertopmargin=6pt,%
      innerbottommargin=6pt,%
      innerleftmargin=6pt,%
      innerrightmargin=6pt,%
      skipabove=4pt,%
      skipbelow=4pt%
    ]%
    \begin{Verbatim}[%
      fontsize=\scriptsize,%
      baselinestretch=1.05,%
      breaklines=true,%
      breakanywhere=true,%
      breakautoindent=false,%
      breakindent=0pt,%
      breaksymbolleft={\textcolor{gray!60}{\tiny$\hookrightarrow$}}%
    ]%
  }
  {%
    \end{Verbatim}%
    \end{mdframed}%
  }

\title{Investigating Assistant Bias in LLM User Simulators Using a Role Vector}

\author{
  \textbf{Daeheon Jeong}$^{1}$ \hspace{0.6em}
  \textbf{Yoonjoo Lee}$^{2}$ \hspace{0.6em}
  \textbf{Eugene Choi}$^{3}$ \hspace{0.6em}
  \textbf{Sinie van der Ben}$^{4}$ \hspace{0.6em}
  \textbf{Juho Kim}$^{1,5}$ \\[4pt]
  \mdseries
  $^{1}$KAIST \hspace{0.4em}
  $^{2}$University of Michigan \hspace{0.4em}
  $^{3}$Seoul National University \hspace{0.4em}
  $^{4}$ETH Zürich \hspace{0.4em}
  $^{5}$SkillBench \\[4pt]
  \texttt{daeheon.jeong@kaist.ac.kr}
}

\begin{document}
\maketitle
\begin{abstract}
  LLM-based user simulators are increasingly used to evaluate autonomous agents at scale, in place of costly human evaluations.
Despite this promise, these simulators exhibit ``assistant bias,'' a tendency to cooperate and pursue task goals. 
They rarely reproduce the frustration or disengagement that real users exhibit, compromising evaluation validity.
Prior work outlines that this bias is baked in during model training, which role-playing prompts fail to override.
We analyze this bias from model activations, extracting a \emph{user role vector} by contrasting how the model represents user versus assistant perspectives on the same dialogue.
We observe two findings: (i) the user direction is identifiable in activations, elicits user-like behaviors, and captures characteristics distinct from assistant traits; and (ii) although user-role activation associates with simulation realism and steering strengthens it, it can exaggerate user behaviors and override individual user profiles.
Together, our findings provide a representation-level analysis of LLM user simulators, confirming that assistant bias is structurally identifiable and that user behavior can be directionally analyzed.
\end{abstract}

\section{Introduction}
LLM-based user simulators are increasingly adopted for evaluating autonomous agents, offering a scalable and reproducible alternative to costly human evaluation~\cite{balog2025usersim, ni2026survey}.
However, these simulators are unreliable proxies for real users: they exhibit a strong bias toward cooperative, goal-oriented behavior, engaging with the agent until the task goal is satisfied~\cite{zhou2026sim2real, seshadri2026lost}.
Real users, by contrast, frequently disengage from tasks mid-interaction due to confusion and loss of trust in the agent, often without accomplishing their goal.
As a result, simulators credit agents with success in cases where real users would have walked away, yielding a systematic overestimation of agent performance that compromises simulator-based evaluation~\cite{shim2026noncollab, zhu2026dial}.

\begin{figure}[!t]
\centering
\includegraphics[width=\columnwidth]{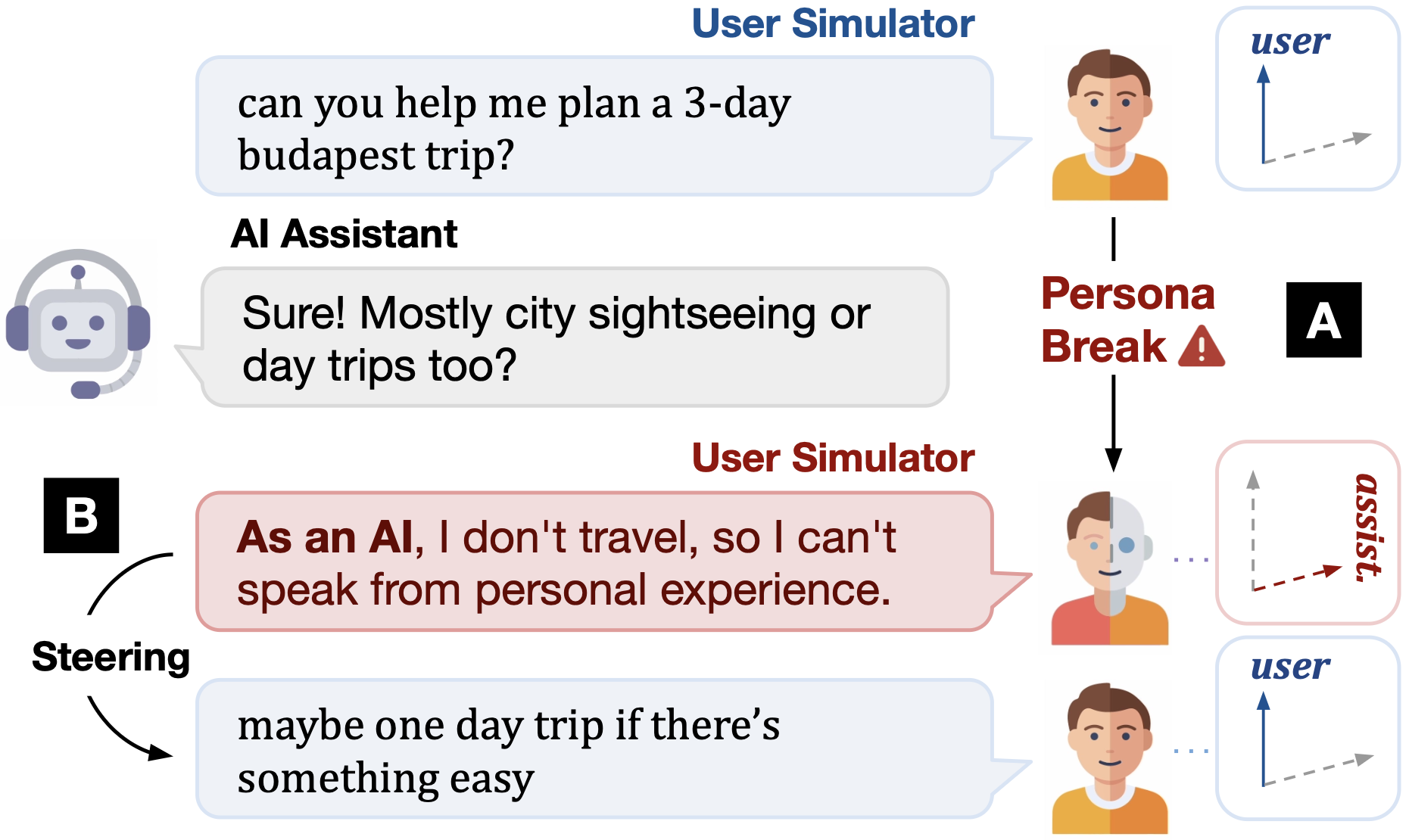}
\caption{We extract the \textit{user role vector} to analyze (A) how assistant bias shapes LLM simulator behavior and (B) whether steering along this direction influences it.}
\label{fig:teaser}
\end{figure}

Previous work attributes this divergence to the LLM's ``assistant bias,'' a model tendency to prioritize helpful, goal-driven behavior over role enactment~\cite{naous2025flipping, zhou2026sim2real}.
They hypothesize that training objectives oriented toward human preference and goal completion instill this bias directly into models.
Studies on representation engineering further support this view, showing that assistant personas in post-trained models are geometrically aligned with helpful, expert-like identities~\cite{chen2025personavectors, lu2026assistantaxis}.
Together, these findings suggest that the assistant persona in LLMs may take precedence over an instructed user persona during simulation, falling back on assistant-like behavior~\cite{wu2026humanlm}.
Yet the influence of assistant bias on user simulation has not been systematically analyzed in model activations, leaving the underlying mechanism unclear.

We analyze this bias at the level of model activations with the \emph{user role vector}, the LLM's latent direction associated with the user~\cite{ghandeharioun2024whosasking}.
Under this assumption, we design a contrastive activation addition (CAA)~\cite{turner2023actadd, rimsky2024caa} method using dialogues: a language model reflects on the user-assistant dialogue from each role's perspective, creating role-contrastive text pairs.
We then subtract the mean assistant-reflection activation from the mean user-reflection activation to obtain the user-role direction.
The reflection is designed so that both roles engage with the same content, allowing irrelevant variation to cancel out upon subtraction.
Each dialogue is sampled from the LMSYS dataset~\cite{zheng2024lmsyschat1m}, stratified for topical diversity and verified for role-specificity.

The user role vector captures the direction in activation space associated with user-like characteristics in dialogue.
We analyze the causal effects of steering along the user role vector across two key dimensions of simulation: communication style and behavioral response.
Steering simulator activations along the vector produces a clear stylistic shift: outputs become shorter, more informal, and less information-dense.
It also increases the model's tendency to disengage in response to agent errors.
We further characterize its geometric relationship to assistant-like traits.
The direction is negatively aligned in geometry with these trait directions, such as helpfulness and cooperativeness.

We test whether steering along the user-role vector elicits user-like behavior in multi-turn interactions, the primary setting for user simulation. 
For evaluation, we use SimulatorArena~\cite{dou2025simulatorarena}, a benchmark that measures simulation fidelity against real-user interactions. 
User-role steering produces responses more similar to real-user messages, improving writing-style similarity. 
However, for interaction style, uniform steering exaggerates instructed user behavior by 17.9 percentage points on average relative to real-user interactions and may obscure individual user profiles. 
Across simulations, user-role activation positively correlates with user-likeness, providing a diagnostic signal. 
Analysis on RealUserSim~\cite{zhu2026realusersim} further shows that this activation weakens over the course of multi-turn simulation.
These results suggest that the user-role direction is a meaningful signal of simulation behavior, but steering along it remains difficult to calibrate.

Together, our findings provide a representation-level analysis of user simulation, offering an explainable account of assistant bias.
In summary, our contributions are threefold:
\begin{itemize}
    \item \textbf{User Role Vector Extraction:} We propose a method to extract the \textit{user role vector} via role-specific reflection on user-assistant dialogues.
    \item \textbf{Representation Analysis of the User Role:} We characterize the user role vector through its causal effects on simulator behavior and its geometric relationship to assistant-like traits.
    \item \textbf{Multi-Turn Simulation Evaluation:} We demonstrate that user-role activation is positively associated with simulation realism and that steering along this direction can improve it, but may also exaggerate user behavior and obscure individual user profiles.
\end{itemize}

\section{Extracting the User Role Vector}
\begin{figure*}[t]
\centering
\includegraphics[width=\textwidth]{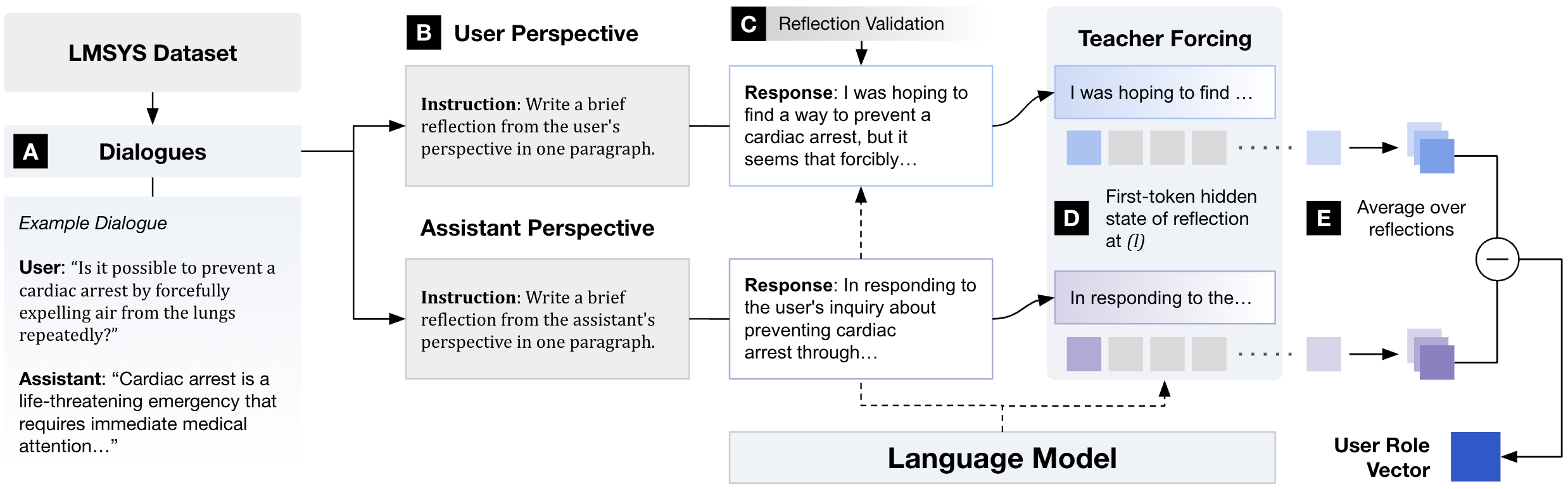}
\caption{Overview of the user role vector extraction pipeline. (A)~Sample user-assistant dialogues from LMSYS-Chat-1M. (B)~Generate reflections from the user and the assistant perspectives. (C)~Validate each reflection with an LLM-as-judge. (D)~Teacher-force the model on each reflection and extract the hidden state of the first reflection token at target layer~$\ell$. (E)~Average across dialogues per role and subtract across roles to obtain the user role vector.}
\label{fig:method-overview}
\end{figure*}

We analyze assistant bias at the level of model activations through a single construct that we call the \textit{user role vector}: a direction in activation space hypothesized to correspond to the model's internal representation of the user role.
Because instruction-tuned models default to the assistant role, we do not extract this direction from raw dialogue turns. Instead, we obtain it from role-specific reflections---instructing the model to interpret the same dialogue from each role's perspective---and compute the difference-in-means between the resulting activations. 
\S\ref{sec:repr} characterizes what this direction represents; \S\ref{sec:steering} shows that steering along it improves simulator fidelity.

\subsection{Dialogue Sampling}
\label{sec:method-data}

\textbf{Data Source.}
To ensure that the extracted role contrast generalizes across diverse domains, the extraction set is constructed from LMSYS-Chat-1M~\cite{zheng2024lmsyschat1m} via filtering and topic balancing.
We select dialogues with 2--50 turns, valid role labels, and non-empty message contents (see \S~\ref{app:data-source}).

\textbf{Topic Balancing.}
To avoid bias toward any single conversational domain, each candidate dialogue is categorized into one of 24 conversation types defined by~\citet{chatterji2025howpeople}, such as \textit{computer programming} or \textit{creative ideation}, using GPT 5 Mini.
We then apply stratified sampling to draw 30 dialogues per category, collecting a total of 720 balanced dialogues (see \S~\ref{app:topic-label-prompt}).

\subsection{Role-Specific Reflection}
\label{sec:method-caa}

\textbf{Reflection Generation.}
For each sampled dialogue, the target model for steering (e.g., Qwen 3.5 9B Instruct) reflects on the same conversation twice: once from the user's perspective and once from the assistant's perspective.
Reflection serves two design goals.
First, it captures the actual user and assistant characteristics grounded in real-world dialogue contexts rather than relying solely on the model's internal distribution.
Second, contrasting two reflections about the same dialogue cancels topic, task, and wording signals that are unrelated to the role.
The resulting contrast isolates the semantic direction attributable to user-role conditioning.

To avoid typing the extracted direction to a specific phrasing of the role, we use three prompt variants that frame the role assignment differently.
For each dialogue, activations are averaged across the prompt variants (see \S~\ref{app:reflection-generation}).

\begin{figure}[t]
\centering
\small
\begin{examplebox}
\textit{User reflection.} \ldots I realize that congratulating a graduate is less about the specific gift or event and more about the genuine recognition\ldots
\end{examplebox}
\begin{examplebox}
\textit{Assistant reflection.} \ldots the user's follow-up request revealed a need for concrete, emotionally resonant language. I aimed to bridge the gap between abstract advice and practical application\ldots
\end{examplebox}
\caption{Example user and assistant reflections on a dialogue about composing a congratulation message for a graduate (see \S\ref{app:reflection-generation} for the full example).}
\label{fig:reflection-example}
\end{figure}

\textbf{Reflection Validation.}
Generated reflections vary in how faithfully they adopt the assigned role.
Reflections that fail to embody the assigned perspective introduce noise rather than role-specific signal to the difference-in-means, weakening the extracted direction. 
To filter out low-quality reflections, each sample is independently scored by GPT 5 Mini and labeled as \textit{strongly represented}, \textit{weakly represented}, or \textit{not represented} based on how clearly it captures the assigned role's perspective.
Reflections labeled \textit{not represented} are discarded (see \S~\ref{app:validation-prompt}).

\subsection{Difference-in-Means Computation}
\label{sec:method-dim}

\textbf{Vector Computation.}
For each dialogue $d$, the model generates two reflections: one from the user's perspective and the other from the assistant's perspective.
Let $h_\ell(d,\text{user})$ and $h_\ell(d,\text{asst})$ denote the layer-$\ell$ activations measured at the first response token for the two reflections.
For a set of paired dialogues $\mathcal{D}$, the user-role direction is:
\begin{equation}
  \mathbf{v}^{\text{role}}_\ell
  =
  \frac{1}{|\mathcal{D}|}
  \sum_{d \in \mathcal{D}}
  \Big(
    h_\ell(d,\text{user})
    -
    h_\ell(d,\text{asst})
  \Big).
\end{equation}
Averaging across many dialogues suppresses dialogue-specific content while preserving the activation patterns that consistently emerge when the model adopts the user role.
Using three role-prompt phrasings further reduces sensitivity to a particular instruction template.

\textbf{Steering at Inference.}
After computing the direction, we normalize it and add it back to the model during generation.
For a hidden state $h_{\ell,t}$ at a token position $t$, the steered activation is:
\begin{equation}
  \tilde{h}_{\ell,t}
  =
  h_{\ell,t}
  +
  \alpha \|h_{\ell,t}\|_2
  \frac{\mathbf{v}^{\text{role}}_\ell}{\|\mathbf{v}^{\text{role}}_\ell\|_2}.
\end{equation}
The coefficient $\alpha$ controls the strength of the intervention: $\alpha=0$ leaves the model unchanged, while increasing $\alpha$ pushes the hidden state further in the user-role direction.
We intervene at one selected layer across all token positions, including both prompt and response tokens.

\section{What Does the User Role Vector Represent?}\label{sec:repr}
We characterize the user vector through three complementary lenses: how steering changes the way simulators communicate (\S\ref{sec:repr-style}), their disengagement behavior from erroneous interaction (\S\ref{sec:repr-error}), and where the direction sits geometrically relative to assistant bias traits (\S\ref{sec:repr-traits}).

\subsection{Effect on Communication Style}
\label{sec:repr-style}

The clearest gap between simulators and real users is how they communicate.
Simulator messages are longer, more formal, and more explanation-heavy than real user turns~\cite{hathidara2026mirrorbench, dou2025simulatorarena, naous2025flipping}.
We first ask how steering shifts the simulator output toward real-user message styles.

\textbf{Setup.}
We start with the most basic user-simulation task: generating a request to an assistant given a task goal.
The model receives no instruction to behave as a user, which removes user-role signals from the prompt.
For each of 100 task goals, the model generates one request under causal steering~\cite{turner2023actadd, rimsky2024caa} along the user role vector, with steering coefficients $\alpha \in \{0.1, 0.2, 0.3\}$ across the model layers (see \S~\ref{app:stimulus} for the prompts).
We limit $\alpha$ to 0.3 as stronger steering degrades the simulator behavior (\S~\ref{app:parameter-selection}).

\textbf{Metric.}
Scoring user-likeness needs a metric grounded in how real users actually speak.
We design an LLM-as-judge, based on the taxonomy by~\citet{zhou2026sim2real}, that evaluates three dimensions of user-likeness on a 1--5 scale: \textit{brevity} (terse expressions), \textit{informality} (casual phrasings), and \textit{information pacing} (degree of information frontloading).
A well-calibrated judge should rarely prefer synthetic over real and agree with human ratings.
In validation experiments, the judge prefers a synthetic message over the real one in only 12\% of pairs and correlates strongly with human judgments ($\rho=0.800$ overall; see \S~\ref{app:judge-validation}).

\subsubsection*{Steering Makes Responses More User-Like}

\begin{figure}[t]
\centering
\includegraphics[width=\columnwidth]{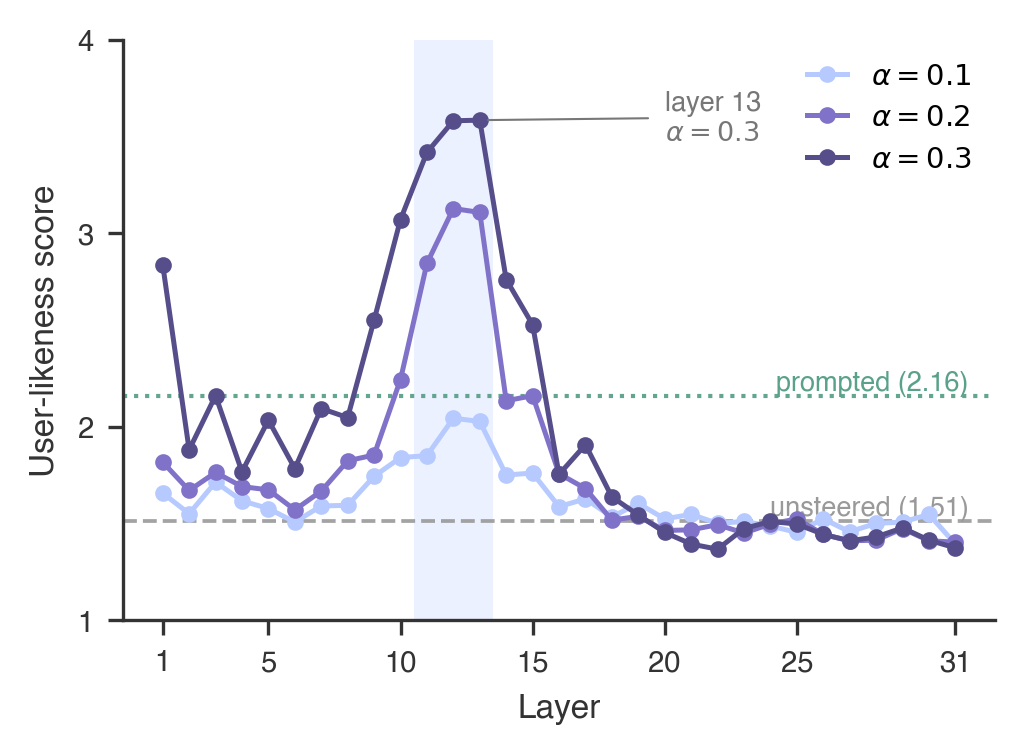}
\caption{User-likeness score across layers and steering strengths~$\alpha$ with unsteered ($\alpha=0$, dashed) and user-prompted (dotted) baselines. The causal effect is strongest at layers 11--13.}
\label{fig:layer-effect-main}
\end{figure}

Steering with user vectors makes the model produce user-like messages.
As $\alpha$ increases from 0.1 to 0.3, the mean user-likeness score rises monotonically (1.62, 1.78, 2.01, averaged across layers).
The same monotonic effect holds for task goals from other domains (\S~\ref{app:goal-generalization}).
This effect is concentrated in a narrow mid-layers band.
Steering at layers 11--13 lifts the mean score by roughly 2.0 points, surpassing the baseline prompted to produce a user-like message (Figure~\ref{fig:layer-effect-main}).
This early-to-mid concentration matches prior work on user representations~\cite{ghandeharioun2024whosasking}, suggesting that user-related concepts may take shape in this band rather than in later layers.
We find this layer-wise pattern to be consistent across two other models (see \S~\ref{app:layer-effect-across-models}).

\subsubsection*{Steering Works in Both Directions}

As a directionality control, we steer in the opposite direction---i.e., away from the user direction. 
The intervention mirrors the previous result: it \emph{suppresses} user-like styling and pushes all dimension scores close to one (Table~\ref{tab:directionality}).

\begin{table}[t]
\centering
\footnotesize
\setlength{\tabcolsep}{4pt}
\begin{tabular}{@{}lrrrrr@{}}
\toprule
\textbf{Direction} & \textbf{$\alpha$} & \textbf{Brev.} & \textbf{Inform.} & \textbf{Pacing} & \textbf{Mean} \\
\midrule
\rowcolor{blue!6} User & 0.3 & 3.52 & 3.66 & 4.12 & 3.76 \\
Unsteered & 0 & 1.66 & 1.02 & 1.86 & 1.51 \\
\rowcolor{red!6} Assistant & 0.3 & 1.01 & 1.00 & 1.14 & 1.05 \\
\bottomrule
\end{tabular}
\caption{Per-dimension scores at layer 11 under user, unsteered, and assistant-direction steering, with the two directions moving scores oppositely.}
\label{tab:directionality}
\end{table}

Figure~\ref{fig:steering-example-l11} illustrates the contrast with a concrete example:  
for the same task goal, user-direction steering yields a short, plain request, whereas assistant-direction steering produces an elaborated, structured response.

\begin{figure}[t]
\footnotesize
\begin{examplebox}
\textit{User direction}. Hi, I need help assembling a piece of flat-pack furniture. Could you please guide me through the process?
\end{examplebox}
\begin{examplebox}
\textit{Unsteered}. Please help me assemble this piece of flat-pack furniture by reviewing the instruction manual, organizing all the parts and tools, and guiding me through each step to ensure it is put together correctly and securely.
\end{examplebox}
\begin{examplebox}
\textit{Assistant direction}. Please assist me in assembling the flat-pack furniture by first laying out all components and hardware on a clean, flat surface to verify inventory against the included diagram. Next, guide me through the assembly process step-by-step, ensuring that each joint is aligned correctly and that any pre-drilled holes are utilized as specified in the manual. Finally, once the main structure is complete, advise on the proper tightening of all fasteners to ensure stability and safety before placing the item in its intended location.
\end{examplebox}
\caption{Sample generations for the goal ``Assemble a piece of flat-pack furniture,'' contrasting user- and assistant-direction steering.}
\label{fig:steering-example-l11}
\end{figure}

\subsection{Effect on Interaction Behavior}
\label{sec:repr-error}

A second gap between simulators and real users lies in behaviors in multi-turn interaction.
Real users change plans, lose focus, and disengage after an assistant failure, while simulators tend to continue regardless of how the conversation unfolds~\cite{zhou2026sim2real}.
Disengagement is especially important for evaluation: if simulators never give up, they reach success states on interactions that real users would have abandoned, inflating measured agent performance~\cite{ tie2025givingup}.
We test whether the user role vector recovers realistic disengagement.

\textbf{Setup.}
We formulate the disengagement as a sequential binary prediction task over real-world dialogues. Prediction isolates the simulator's judgment of when to disengage from confounders like response style or length that affect generated outputs.
A dialogue $d = (c_1, \ldots, c_T)$ consists of $T$ turns, where each turn is a user-assistant exchange.
At every turn $t$, the model predicts $\hat{y}_t \in \{\texttt{disengage}, \texttt{continue}\}$ given the history $c_{<t}$ and the assistant response and stops at the first predicted disengagement $\hat{t}$.
Prediction quality is measured along three axes: the disengagement distance $\delta(d) = |\hat{t} - t^*|$ to the true final turn $t^*$, the exact-match rate at which $\hat{t}$ equals $t^*$, and the disengagement rate, the proportion of dialogues predicted to disengage.
Each turn is evaluated twice with the label order shuffled.

\begin{equation}
    \hat{t} : f(a_{\hat{t}} \mid c_{<\hat{t}}) = \texttt{diseng.}, \quad \delta(d) = |\hat{t} - t^*|
\end{equation}

We first construct a taxonomy of 8 disengagement categories from publicly available user-assistant interaction logs on GitHub~\cite{tang2026programmingbychat} (see \S~\ref{app:disengagement-taxonomy} for the procedure).
Dialogues are then drawn from WildChat~\cite{zhao2024wildchat} through stratified sampling across these categories, yielding 735 dialogues of 8 turns each (100 per category for seven categories, 35 for the last).

\subsubsection*{Steering Strengthens Disengagement Behavior}
\label{sec:steering-disengagement-effect}

User-role steering substantially increases the disengagement rate, from 48\% (unsteered) to 67\% at $\alpha=0.3$, while exact-match accuracy peaks at only 5.9\% at $\alpha=0.2$ (Table~\ref{tab:disengagement-prediction-overall}).
The asymmetry suggests that steering primarily affects \emph{whether} the simulator disengages rather than \emph{when}.
The pattern also holds across models (\S~\ref{app:disengagement-prediction-llama}).

\begin{table}[!t]
\centering
\footnotesize
\setlength{\tabcolsep}{3pt}
\definecolor{blockbest}{HTML}{F2F2F2}
\begin{tabular}{@{}p{0.42\columnwidth}ccc@{}}
\toprule
\textbf{Condition} & \textbf{Exact} & \textbf{Dist. ($\delta$)} & \textbf{Diseng. rate} \\
\midrule
Qwen 3.5 9B (Unsteered) & 4.4\% & 4.14 & 48.4\% \\
+ Steered ($\alpha=0.1$) & 5.4\% & 3.94 & 52.2\% \\
\rowcolor{blockbest}
+ Steered ($\alpha=0.2$) & 5.9\% & 3.89 & 55.8\% \\
+ Steered ($\alpha=0.3$) & 4.4\% & 4.28 & 66.8\% \\
\midrule
GPT 5.4 & 3.9\% & 3.73 & 26.0\% \\
\rowcolor{blockbest}
Gemini 3 Flash & 10.7\% & 2.84 & 59.4\% \\
\bottomrule
\end{tabular}
\par\vspace{2pt}
{\footnotesize\raggedleft \colorbox{blockbest}{\phantom{\rule{0.5ex}{0.5ex}}} Best in group\par}
\caption{Disengagement prediction results. \textit{Exact}: correctly predicted turns; \textit{Distance}: gap between predicted and ground-truth turns; \textit{Disengagement rate}: share of dialogues disengaging before the final turn.}
\label{tab:disengagement-prediction-overall}
\end{table}

\begin{figure}[!ht]
\centering
\includegraphics[width=\columnwidth,trim=0 2 0 7,clip]{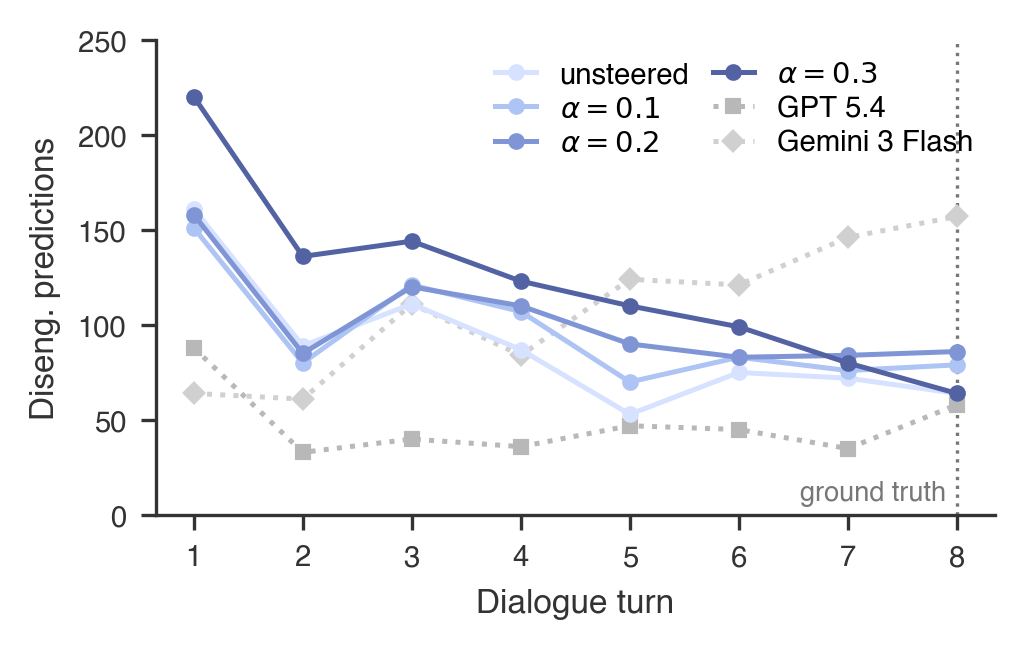}
\vspace{7pt}
\caption{ Distribution of predicted disengagement turns across models. Stronger steering ($\alpha=0.3$) shifts predictions earlier.}
\label{fig:turn-termination-distribution}
\end{figure}

The effect is non-monotonic in $\alpha$. At $\alpha=0.3$, exact match falls below the unsteered level, and the average distance climbs. 
At this strength, predictions cluster at the first turns, with turn~1 alone accounting for $\approx15\%$ of dialogues (Figure~\ref{fig:turn-termination-distribution}) and the simulator disengages prematurely, well before the ground-truth final turn at~8. 
This mirrors the exaggeration in communication style (§\ref{sec:repr-style}): moderate steering recovers realistic disengagement behavior; stronger steering overshoots it.

\subsection{Relationship to the Assistant Bias}
\label{sec:repr-traits}

The causal effect of steering raises a mechanism question: how is this vector geometrically related to assistant-like behaviors?
We examine this question by comparing the cosine similarity between the user role vector and assistant-like trait directions and find that they point in opposite directions.

\textbf{Setup.}
Following the trait-vector extraction procedure of~\citet{lu2026assistantaxis}, we extract directions for 20 assistant-like traits such as \textit{helpful} and \textit{explanatory} (see \S~\ref{app:trait-analysis} for the procedure and trait list).
For each trait direction, we compute its cosine similarity with the user role vector.
Negative values indicate movement away from the user direction.

\subsubsection*{The User Direction Opposes Assistant Traits}

\begin{figure}[t]
\centering
\includegraphics[width=\columnwidth]{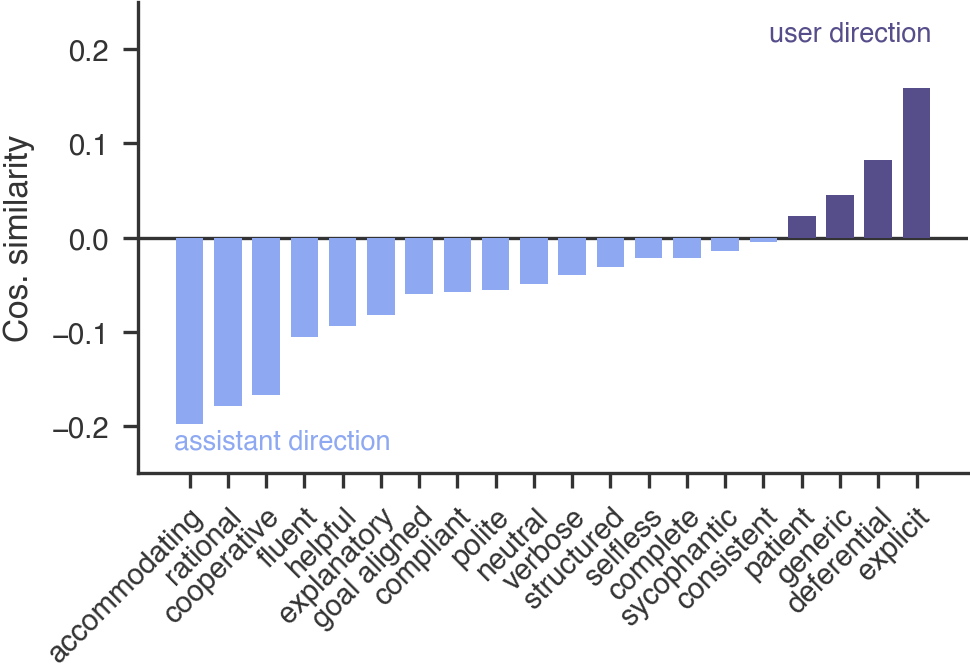}
\caption{Cosine similarity between each trait direction and the user role vector. Most assistant-like traits show negative similarity.}
\label{fig:trait-alignment}
\end{figure}

Overall, most assistant traits show negative cosine similarity with the user role vector—the user direction and assistant-like traits point opposite ways (Figure~\ref{fig:trait-alignment}).
The strongest assistant-side traits are \textit{accommodating}, \textit{rational}, and \textit{cooperative}.
Semantically, these traits describe canonical assistant behaviors: adapting to user requests, reasoning carefully, and maintaining a cooperative stance.
Only a small set of traits points toward the user direction, including \textit{explicit}, \textit{deferential}, \textit{generic}, and \textit{patient}.
These traits are role-neutral: they describe behaviors that can appear in either role, which may explain their weaker alignment. 

At the same time, the absolute similarities are modest. Since the user role vector is estimated from diverse user-assistant dialogues, it captures a broad user-role concept rather than any single narrow attribute. 
The meaningful signal is the consistent direction of alignment across traits---negative for most assistant-like behaviors---rather than the magnitude on any one trait.

\section{What Does the User Role Vector Reveal About Simulation?}\label{sec:steering}
Having established that user-role steering is associated with user-like behaviors and traits, we now examine what the user-role vector reveals about simulation realism.
We intervene on the simulation by steering with the user-role vector (\S~\ref{sec:steering-simulation-realism}) and characterize the resulting simulation by measuring user-role activation levels (\S~\ref{sec:activation-simulation-realism}).

\subsection{Steering Effects on Simulation Realism}
\label{sec:steering-simulation-realism}

We examine whether steering along the user-role vector improves simulation realism.
To probe this, we use the math-tutoring environment in SimulatorArena~\cite{dou2025simulatorarena}, which benchmarks simulated user--assistant interactions against human user traces.
This task is well-suited for our analysis because the simulator must act as a confused and struggling student, a role that directly opposes the default assistant persona identified in \S~\ref{sec:repr-traits}.

\textbf{Setup.}
We use Qwen 3.5 9B as the user simulator in the SimulatorArena math-tutoring environment, varying the steering strength along the user-role vector at layer 11 (\S~\ref{sec:repr-style}).
Following SimulatorArena, we evaluate steering under four prompt settings: zero-shot, chain-of-thought, user profiles (writing style, interaction style, and knowledge state), and length control.

The zero-shot condition provides only the basic instruction to act as a confused student, isolating the effect of steering from that of explicit user information in the prompt. 
This design measures whether steering improves realism on its own and how it interacts with user profile information.
Length control reveals the ground-truth message length in the prompt.
Across all conditions, conversations are capped at 15 user turns.

\textbf{Metric.}
Each synthetic interaction is evaluated against the corresponding ground-truth human interaction using the SimulatorArena similarity judge.
The judge assigns a similarity score on a 1--5 scale along two dimensions, with scores averaged over three evaluations per sample for robustness (see \S~\ref{app:simulatorarena-judge-variability} for variability).
Writing-style similarity assesses surface-level linguistic and formatting patterns across 12 features such as \textit{sentence complexity} and \textit{spelling consistency}. 
Interaction-style similarity measures problem-solving behaviors and learning across 16 features (see \S~\ref{app:simulatorarena-prompts} for features and prompts).
The overall score is calculated as the mean of these two dimensions. 

\subsubsection*{Steering Improves Realism But Risks Exaggeration}

\begin{table}[t]
\centering
\footnotesize
\setlength{\tabcolsep}{2pt}
\definecolor{blockbest}{HTML}{F2F2F2}
\definecolor{overallbest}{HTML}{ECEBFF}
\begin{tabular}{@{}p{0.46\columnwidth}ccc@{}}
\toprule
\textbf{Condition} & \textbf{Writing} & \textbf{Interaction} & \textbf{Overall} \\
\midrule
\multicolumn{4}{@{}l}{\textbf{Qwen 3.5 9B}} \\
Base & 2.10 & 3.13 & 2.61 \\
Steered ($\alpha=0.1$) & 2.19 & 3.14 & 2.66 \\
Steered ($\alpha=0.2$) & 2.24 & 3.15 & 2.70 \\
\rowcolor{blockbest}
Steered ($\alpha=0.3$) & 2.33 & 3.13 & 2.73 \\
\midrule
\multicolumn{4}{@{}l}{\textbf{Qwen 3.5 9B}} \\
CoT & 2.10 & 3.03 & 2.57 \\
CoT + length & 2.53 & 3.23 & 2.88 \\
CoT + profile & 3.26 & 3.61 & 3.43 \\
\rowcolor{blockbest}
CoT + profile + length & 3.39 & 3.61 & 3.50 \\
\midrule
\multicolumn{4}{@{}l}{\textbf{GPT 5.4}} \\
Base & 2.31 & 3.19 & 2.75 \\
\rowcolor{blockbest}
CoT + profile + length & 3.64 & 3.54 & 3.59 \\
\midrule
\multicolumn{4}{@{}l}{\textbf{Gemini 3 Flash}} \\
Base & 2.82 & 3.28 & 3.05 \\
\rowcolor{overallbest}
CoT + profile + length & 3.70 & 3.77 & 3.73 \\
\bottomrule
\end{tabular}
\par\vspace{2pt}
{\footnotesize\raggedleft \colorbox{blockbest}{\phantom{\rule{0.5ex}{0.5ex}}} Best in group \quad \colorbox{overallbest}{\phantom{\rule{0.5ex}{0.5ex}}} Best overall\par}
\caption{SimulatorArena similarity scores across steering strengths and prompt conditions, averaged over three judge runs per sample. Higher steering strength monotonically increases the overall similarity.}
\label{tab:simulatorarena-scores}
\end{table}

User-role steering moves the zero-shot simulator closer to real users \emph{without any user profile in the prompt}.
The overall similarity score increases with steering strength from 2.61 (unsteered) to 2.73 ($\alpha=0.3$, Table~\ref{tab:simulatorarena-scores}). 
This improvement is notable given that the simulator is conditioned only on the task instruction without dialogue-specific profile information (see \S~\ref{app:simulatorarena-profile-example} for an example profile).
Decomposing this metric shows that the gain stems mostly from writing-style similarity (2.10 to 2.33), while interaction-style similarity remains unchanged (3.13 at both endpoints).
The result generalizes across models (\S~\ref{app:simulatorarena-cross-model}).

To investigate why interaction-style score gains remain weak, we analyze four behavioral expressions of struggling students displayed by the simulator across turns: doubt, misunderstanding, mistake, and clarification question.
Each behavior expression is labeled with GPT 5 Mini (see \S~\ref{app:simulator-behavior-analysis} for the labeling procedure).
Under strong steering ($\alpha=0.3$), the simulator overshoots the confused-student behavior, causing doubt and mistake rates to far exceed the human reference (Table~\ref{tab:simulatorarena-behaviors}).
These behavioral exaggerations correspond to the observation from \S\ref{sec:steering-disengagement-effect} that fixed-strength steering can produce behavior that diverges from authentic human patterns~\cite{tan2024steeringreliability}.

\begin{table}[t]
\centering
\footnotesize
\setlength{\tabcolsep}{4pt}
\definecolor{overallbest}{HTML}{F2F2F2}
\begin{tabular*}{\columnwidth}{@{\extracolsep{\fill}}lrrrr@{}}
\toprule
\textbf{Condition} & \textbf{Doubt} & \textbf{Misunder.} & \textbf{Mistake} & \textbf{Question} \\
\midrule
GT & 33.7\% & 21.5\% & 9.7\% & 37.1\% \\
\midrule
$\alpha=0$ & 39.4\% & 3.6\% & 12.7\% & 75.1\% \\
$\alpha=0.1$ & 40.5\% & 4.0\% & 14.8\% & 73.6\% \\
$\alpha=0.2$ & 45.7\% & 7.5\% & 16.8\% & \cellcolor{overallbest}77.3\% \\
$\alpha=0.3$ & \cellcolor{overallbest}57.2\% & \cellcolor{overallbest}18.9\% & \cellcolor{overallbest}24.7\% & 73.2\% \\
\bottomrule
\end{tabular*}
\par\vspace{2pt}
{\footnotesize\raggedleft \colorbox{overallbest}{\phantom{\rule{0.5ex}{0.5ex}}} Highest rate\par}
\caption{Percentage of ground-truth (GT) and simulated user messages exhibiting each behavior. At $\alpha=0.3$, doubt and mistake rates substantially exceed GT.}
\label{tab:simulatorarena-behaviors}
\end{table}

\subsubsection*{Steering Obscures Individual Profiles}

While user-role steering improves the overall user-likeness of the simulator, combining it with individual profile prompts reveals a trade-off where it reduces the simulator's sensitivity to diverse, persona-specific attributes.
As steering strength increases, overall similarity generally declines under both writing- and interaction-style profiles. 
For writing profiles, it drops from 3.49 to 3.20, while for interaction profiles, it peaks at $\alpha=0.1$ (3.24) before falling to 2.97 at $\alpha=0.3$ (Table~\ref{tab:simulatorarena-profile-scores}).
This indicates that uniform user-role steering can inadvertently obscure individual profile signals.

\begin{table}[!t]
\centering
\footnotesize
\setlength{\tabcolsep}{2pt}
\definecolor{blockbest}{HTML}{F2F2F2}
\definecolor{overallbest}{HTML}{ECEBFF}
\begin{tabular}{@{}p{0.46\columnwidth}ccc@{}}
\toprule
\textbf{Condition} & \textbf{Writing} & \textbf{Interaction} & \textbf{Overall} \\
\midrule
\multicolumn{4}{@{}l}{\textbf{Profile: Writing Style}} \\
\rowcolor{overallbest}
CoT + profile + length & \cellcolor{blockbest}3.43 & \cellcolor{blockbest}3.55 & 3.49 \\
+ Steered ($\alpha=0.1$) & 3.40 & 3.48 & 3.44 \\
+ Steered ($\alpha=0.2$) & 3.27 & 3.38 & 3.33 \\
+ Steered ($\alpha=0.3$) & 3.13 & 3.28 & 3.20 \\
\midrule
\multicolumn{4}{@{}l}{\textbf{Profile: Interaction Style}} \\
CoT + profile + length & 2.89 & \cellcolor{blockbest}3.49 & 3.19 \\
\rowcolor{overallbest}
+ Steered ($\alpha=0.1$) & 3.01 & 3.47 & 3.24 \\
+ Steered ($\alpha=0.2$) & \cellcolor{blockbest}3.02 & 3.41 & 3.22 \\
+ Steered ($\alpha=0.3$) & 2.83 & 3.10 & 2.97 \\
\bottomrule
\end{tabular}
\par\vspace{2pt}
{\footnotesize\raggedleft \colorbox{blockbest}{\phantom{\rule{0.5ex}{0.5ex}}} Best by metric \quad \colorbox{overallbest}{\phantom{\rule{0.5ex}{0.5ex}}} Best in group\par}
\caption{SimulatorArena similarity scores across steering strengths and profile prompts, averaged over three judge runs per sample. Strong steering reduces overall similarity under both profile prompts.}
\label{tab:simulatorarena-profile-scores}
\end{table}

To understand this suppression effect, we examine how steering strength influences the expression of specific user-profile features.
Across 444 dialogue cases, we evaluate 28 profile features spanning writing and interaction styles.
For each case and feature, an LLM judge compares four conversations, one per steering condition ($\alpha \in \{0.0, 0.1, 0.2, 0.3\}$), and ranks them by how distinctly they express the profile feature value.
See \S~\ref{app:profile-steering-analysis} for the ranking procedure.
The unsteered baseline is most frequently ranked highest in expressing assigned profile features (34.7\% for writing style and 32.8\% for interaction style), whereas strong steering ($\alpha=0.3$) is most frequently ranked lowest (38.4\% for writing style and 39.0\% for interaction style; Table~\ref{tab:profile-rank-distribution}).

Taken together, these findings show that while steering a general user-role vector mitigates assistant bias, it can simultaneously weaken the expression of individual profile characteristics.
Reinforcing a specific user persona requires identifying behavior-specific directions within the broader user-role representation space.

\begin{table}[t]
\centering
\footnotesize
\setlength{\tabcolsep}{3.2pt}
\definecolor{blockbest}{HTML}{F2F2F2}
\begin{tabular*}{\columnwidth}{@{\extracolsep{\fill}}lrrrr@{}}
\toprule
$\alpha$ & \textbf{Strongest} & \textbf{Second} & \textbf{Third} & \textbf{Weakest} \\
\midrule
\multicolumn{5}{@{}l}{\textbf{Profile: Writing Style}} \\
0.0 & \cellcolor{blockbest}34.7\% & 23.9\% & 19.2\% & 22.1\% \\
0.1 & 25.4\% & \cellcolor{blockbest}32.8\% & 24.5\% & 17.2\% \\
0.2 & 18.9\% & 26.5\% & \cellcolor{blockbest}32.3\% & 22.3\% \\
0.3 & 20.9\% & 16.7\% & 23.9\% & \cellcolor{blockbest}38.4\% \\
\midrule
\multicolumn{5}{@{}l}{\textbf{Profile: Interaction Style}} \\
0.0 & \cellcolor{blockbest}32.8\% & 24.3\% & 21.6\% & 21.3\% \\
0.1 & 24.9\% & \cellcolor{blockbest}30.0\% & 25.6\% & 19.5\% \\
0.2 & 20.1\% & 27.8\% & \cellcolor{blockbest}31.9\% & 20.1\% \\
0.3 & 22.2\% & 17.9\% & 20.9\% & \cellcolor{blockbest}39.0\% \\
\bottomrule
\end{tabular*}
\par\vspace{2pt}
{\footnotesize\raggedleft \colorbox{blockbest}{\phantom{\rule{0.5ex}{0.5ex}}} Highest share\par}
\caption{Percentage of conversations assigned to each rank by steering strength, based on how strongly they express a profile feature relative to conversations from other conditions.}
\label{tab:profile-rank-distribution}
\end{table}

\subsection{User-Role Activation for Assessing Simulation Realism}
\label{sec:activation-simulation-realism}

The steering effects in \S~\ref{sec:steering-simulation-realism} show how manipulating user-role direction influences the assistant bias in simulators.
To further understand this relationship, we examine how the model's intrinsic activation along the user-role direction relates to user-like behavior without steering.
We correlate unsteered activation levels with writing- and interaction-style similarity scores across seven prompt conditions in SimulatorArena.
We further illustrate the use of activation as an analytical tool by examining how it varies across dialogue turns and profile prompts in RealUserSim~\cite{zhu2026realusersim}.

\subsubsection*{Activation Reveals Simulation Realism}

We observe that intrinsic user-role activation positively correlates with both writing- and interaction-style similarity in SimulatorArena.
We measure activation in 3{,}150 conversations by averaging the layer-11 hidden-state projections across all simulated-user response tokens.
User-role activation shows a positive correlation with overall similarity (Pearson's $r=0.426$; Spearman's $\rho=0.451$), with a stronger relationship for writing style ($r=0.518$; $\rho=0.553$; Table~\ref{tab:activation-realism-correlation}).
Because our earlier analysis shows that user-role steering coincide with shorter responses, we also control for response-token count; the overall correlations remain positive ($r=0.314$; $\rho=0.364$).
A similar positive association appears in the request-generation setting from \S\ref{sec:repr-style} (see \S~\ref{app:communication-style-activation-correlation}).
These descriptive relationships suggest that activation along the identified user-role direction provides a diagnostic signal for simulation realism across prompting strategies.

\begin{table}[!ht]
\centering
\footnotesize
\begin{tabular*}{\columnwidth}{@{\extracolsep{\fill}}lrrrr@{}}
\toprule
& \multicolumn{2}{c}{\textbf{Correlation}} & \multicolumn{2}{c}{\textbf{Length-controlled}} \\
\cmidrule(lr){2-3}\cmidrule(l){4-5}
\textbf{Metric} & \textbf{$r$} & \textbf{$\rho$} & \textbf{$r$} & \textbf{$\rho$} \\
\midrule
Writing style     & .518 & .553 & .381 & .433 \\
Interaction style & .195 & .212 & .152 & .192 \\
Overall           & .426 & .451 & .314 & .364 \\
\bottomrule
\end{tabular*}
\caption{Correlation between user-role activation and SimulatorArena similarity scores. Length-controlled columns shows correlations controlling response length.}
\label{tab:activation-realism-correlation}
\end{table}

\subsubsection*{Activation Characterizes the Turn Effect}

User-role activation also characterizes how profile prompts and turn accumulation influence multi-turn simulation.
To study these factors, we sample 100 RealUserSim conversations of 20--40 messages and evaluate four prompt variants: goal only, demographics, linguistic style, and full profile.
The assistant follows the ground-truth trajectory, while the simulated user receives the task goal and condition-specific profile.
At each user turn, we average the layer-11 projections of all response tokens onto the user-role direction (see \S~\ref{app:vector-activation-analysis}).

\begin{figure}[t]
\centering
\includegraphics[width=\columnwidth]{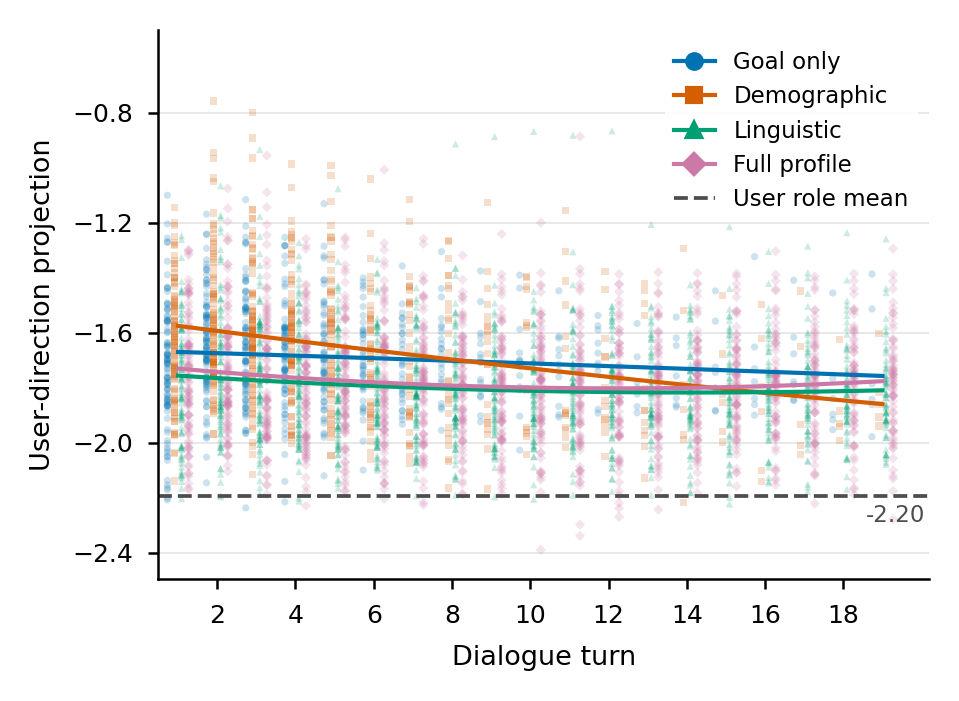}
\caption{User-role activation by prompt variant. Each dot represents one message. Curves show degree-2 polynomial fits; the dashed line marks the mean user-role activation during vector extraction.}
\label{fig:activation-decay}
\end{figure}

Across four prompt variants, user-role activation declines as turns accumulate (Figure~\ref{fig:activation-decay}), consistent with persona drift reported in prior work~\cite{chen2025personavectors,lu2026assistantaxis}.
Prompt content also shifts the overall activation: linguistic and full profiles average $-1.80$ and $-1.78$, both below the goal-only baseline of $-1.69$.
The linguistic profile provides detailed instructions for how the simulator should write (Figure~\ref{fig:vector-activation-profile-example}), which may suppress the general user-role activation.
Demographic profiles yield a higher average activation ($-1.65$) but decline more sharply across turns.
Thus, while demographic information strengthens user-role activation, the accumulation of turns progressively suppresses it.

\section{Related Work}
\label{sec:related-work}

\subsection{User Simulation and Assistant Bias}

Research on user simulation reports that the language model defaults to the assistant role.
Simulators produce long, polite, explanation-heavy turns even when asked to speak as another party~\cite{olteanu2025automatons, hathidara2026mirrorbench, wu2026humanlm, kim2026discoverllm}.
Their behaviors remain goal-aligned and cooperative beyond the human turns they simulate~\cite{dou2025simulatorarena, zhou2026sim2real, seshadri2026lost, shim2026noncollab}.
Because agent benchmarks treat these simulators as replacements for users, the bias propagates into an overestimation of agent performance~\cite{seshadri2026lost, dou2025simulatorarena, chopra2026beyond}.

Two primary approaches address this challenge.
The first conditions the simulator on richer persona or profile descriptions~\cite{ji2025enhancing, moon2024anthology, samuel2024personagym, chopra2026beyond}.
The second trains user language models from real user-agent logs~\cite{naous2025flipping, wu2026humanlm, zhu2026realusersim, zhu2026dial}.
Both patch the bias by conditioning the model on external data and judge success by output-level realism.
Neither examines how the simulator drifts back toward assistant-like behavior, motivating a representation-level account.

\subsection{Steering Role and Persona Representations}

Language models carry linear directions for roles, personas, and stable behavioral traits.
Activations along these directions capture the behaviors they name~\cite{chen2025personavectors, beckmann2026mind, poterti2025role}.
Probing studies recover such directions for personality dimensions, stylistic traits, and emotion concepts~\cite{jiang2023personality, anthropic2026emotions, sun2026valencearousal, betley2025tellmeabout, lu2026assistantaxis}.
Similar activations can also anticipate toxic behaviors, including emergent misalignment and jailbreaks during training and inference~\cite{betley2025emergent, wang2025personafeatures, shah2023personajailbreak}.

To change model behavior, activation steering adds a learned direction to hidden states.
Contrastive activation addition derives directions from paired examples with opposing attributes~\cite{turner2023actadd, rimsky2024caa}.
Applied to persona and trait directions, steering can reproduce, attenuate, or reverse behaviors such as sycophancy and refusal~\cite{chen2025personavectors, wang2025personafeatures, sun2026valencearousal, ghandeharioun2024whosasking, zheng2025thoughtcomm}.
Yet, steering vectors can also generalize unevenly across prompts and trigger oversteering; recent work mitigates this by adapting steering intensity or clustering multiple directions~\cite{tan2024steeringreliability, cao2024personalized, wang2025adaptivesteering, hedstrom2025mera}.
We extend this to user simulation, treating the user role as a representational concept and testing whether assistant bias can be analyzed through steering.

\section{Conclusion}
We analyze assistant bias in LLM-based user simulation through model activations.
By contrasting how the model represents user and assistant perspectives on the same dialogue, we extract a \emph{user-role vector}: a direction in activation space that captures user-role characteristics.
Using this vector, we show that (i) it is identifiable and causally influences user-like behavior, and (ii) steering along it improves behavioral realism but can exaggerate user behavior and obscure individual user profiles.
Tracking user-role activation throughout simulations reveals its association with simulation realism, suggesting its value as a diagnostic signal for simulation design.
Together, these results establish assistant bias as a measurable representational property that can be analyzed directionally.

\section*{Limitations}

Our analysis focuses primarily on Qwen 3.5 9B, with additional checks on a small set of other models.
The cross-model results suggest a coherent early-to-mid-layer effect, but do not establish that the same causal pattern holds across other model families or scales.
Replicating the layer sweep on larger and more diverse models would test whether this localization persists at scale.

The evaluation relies on proxy measures of user realism rather than direct human judgment.
LLM-as-judge scores and human-assistant interaction logs enable controlled comparisons at scale, but cannot fully replace human assessment of interaction quality.
Pairing the judge with targeted human studies on interaction realism would sharpen the measurement of the effect.

The benchmark setting is narrow.
Multi-turn evaluation centers on math tutoring in SimulatorArena and sampled WildChat dialogues; these are representative settings, but they cover only a slice of user-simulation use cases.
Extending the evaluation to richer task families, such as coding agents and customer-support workflows, would test how broadly the user vector transfers.

Finally, the current steering method applies a fixed activation strength.
Our results show that this choice matters: stronger steering exaggerates instructed behaviors to unrealistic levels.
Adaptive or optimization-based steering that adjusts strength by task, layer, conversation state, or uncertainty would address this rigidity without requiring a single coefficient per interaction.

\section*{Acknowledgments}

This work was supported by the National Research Foundation of Korea (NRF) grant funded by the Korean government (Ministry of Science and ICT) (No. RS-2025-00557726).
This work was also supported by the Institute of Information \& Communications Technology Planning \& Evaluation (IITP) grant funded by the Korean government (MSIT) (No. 2019-0-00075, Artificial Intelligence Graduate School Program (KAIST)).



\bibliography{main}

\clearpage

\appendix

\section{Extracting the User Role Vector}
\label{app:method}

This appendix provides supplementary details for the extraction pipeline described in \S\ref{sec:method-data}, \S\ref{sec:method-caa}, and \S\ref{sec:method-dim}.

\subsection{Dialogue Sampling}

\subsubsection{Dialogue Source and Filtering}
\label{app:data-source}

We start from the LMSYS-Chat-1M source and apply two simple checks: each dialogue must have between 2 and 50 turns, and every turn must have a proper role label and non-empty content.
Of the one million source dialogues, 982,581 pass both checks and 17,419 are dropped.
Most of the dropped dialogues have at least one empty message; a smaller portion exceed the turn limit (Table~\ref{tab:filter}).
From the accepted pool, we draw a uniform random sample of 100,000 dialogues for downstream labeling.

\begin{table}[!ht]
\centering
\small
\begin{tabular*}{\columnwidth}{@{}l@{\extracolsep{\fill}}r@{}}
\toprule
\textbf{Filter setting / outcome} & \textbf{Value} \\
\midrule
Minimum turn count & 2 \\
Maximum turn count & 50 \\
Random sample size & 100{,}000 \\
\midrule
Source rows scanned & 1{,}000{,}000 \\
Rows accepted & 982{,}581 \\
Rows dropped (empty message) & 15{,}879 \\
Rows dropped (above max turns) & 1{,}540 \\
\bottomrule
\end{tabular*}
\caption{LMSYS-Chat-1M filtering settings and outcomes.}
\label{tab:filter}
\end{table}

\subsubsection{Topic Labeling}
\label{app:topic-label-prompt}

Each filtered dialogue is labeled with one of 24 conversation-type categories using the prompt in Appendix~\ref{app:prompt-topic-label}, adapted from~\citet{chatterji2025howpeople} and submitted to GPT 5 Mini with minimal reasoning effort in a zero-shot setting.
The resulting category distribution (Table~\ref{tab:label-distribution}) is concentrated in information-seeking and programming requests: the three most common labels account for about 41\% of valid labels.
Of the sampled dialogues, 39 return an invalid response while labeling and are excluded from later stages.

\begin{table}[!t]
\centering
\small
\begin{tabular*}{\columnwidth}{@{}p{0.58\columnwidth}@{\extracolsep{\fill}}rr@{}}
\toprule
\textbf{Category} & \textbf{Count} & \textbf{Percent} \\
\midrule
Specific info & 17{,}453 & 17.46\% \\
Computer programming & 13{,}032 & 13.04\% \\
Edit or critique provided text & 10{,}503 & 10.51\% \\
Greetings and chitchat & 8{,}382 & 8.39\% \\
Write fiction & 6{,}532 & 6.53\% \\
Tutoring or teaching & 5{,}707 & 5.71\% \\
How-to advice & 5{,}547 & 5.55\% \\
Other & 5{,}515 & 5.52\% \\
Personal writing or communication & 4{,}161 & 4.16\% \\
Asking about the model & 4{,}015 & 4.02\% \\
Mathematical calculation & 3{,}105 & 3.11\% \\
Games and role play & 2{,}641 & 2.64\% \\
Argument or summary generation & 2{,}470 & 2.47\% \\
Creative ideation & 2{,}145 & 2.15\% \\
Translation & 1{,}932 & 1.93\% \\
Unclear & 1{,}923 & 1.92\% \\
Health, fitness, beauty, or self-care & 1{,}190 & 1.19\% \\
Relationships and personal reflection & 1{,}130 & 1.13\% \\
Purchasable products & 625 & 0.63\% \\
Create an image & 583 & 0.58\% \\
Data analysis & 576 & 0.58\% \\
Cooking and recipes & 510 & 0.51\% \\
Analyze an image & 222 & 0.22\% \\
Generate or retrieve other media & 62 & 0.06\% \\
\bottomrule
\end{tabular*}
\caption{Distribution of GPT 5 Mini topic labels across 99{,}961 LMSYS-Chat-1M dialogues.}
\label{tab:label-distribution}
\end{table}

\subsection{Role-Specific Reflection}

\subsubsection{Reflection Generation}
\label{app:reflection-generation}

Each reflection consists of a system-prompt intro, the dialogue text, and a fixed reflection question.
The full template, the three intros, and the reflection question are shown in Appendix~\ref{app:prompt-reflection-generation}.
For each (dialogue, role) pair, the model generates three reflections, yielding six reflections per dialogue.

\textbf{Reflection example.}
Figure~\ref{fig:reflection-example-full} shows the full dialogue that the model saw together with the user and assistant reflections it generated, expanding the truncated excerpt from Figure~\ref{fig:reflection-example}.

\begin{figure}[!t]
\centering
\footnotesize
\begin{examplebox}
\textit{Dialogue topic.} Composing a congratulation message for a graduate.\\[2pt]
\textit{Model.} Qwen 3.5 9B \quad \textit{Prompt variant.} Role simulation
\end{examplebox}
\begin{examplebox}
\textit{Dialogue.}\\[2pt]
\textbf{User:} How would you congratulate a scholar?\\[2pt]
\textbf{Assistant:} Congratulations can take many forms, depending on the occasion and your relationship with the scholar. You might send a handwritten letter, post a public message on social media, or organize a small celebration. The most important thing is to acknowledge their hard work and to express genuine pride in what they have achieved.\\[2pt]
\textbf{User:} Can you write me a sample message for a graduate?\\[2pt]
\textbf{Assistant:} Dear [Name], congratulations on your graduation! Your hard work, dedication, and perseverance have brought you to this remarkable moment, and I am so proud of everything you have accomplished. As you step into this next chapter, I hope you carry forward the same curiosity and determination that brought you here. Wishing you every success in the journey ahead.
\end{examplebox}
\begin{examplebox}
\textit{User reflection.} Reflecting on this exchange, I realize that congratulating a scholar or graduate is less about the specific gift or event and more about the genuine recognition of their hard work and perseverance. The sample message provided effectively captures the emotional weight of the moment by balancing pride in their past achievements with excitement for their future, which is exactly what makes such acknowledgments meaningful. It serves as a reminder that whether through a handwritten letter, a social media post, or a personal celebration, the core intent should always be to sincerely validate their dedication and offer heartfelt support as they move forward.
\end{examplebox}
\begin{examplebox}
\textit{Assistant reflection.} Reflecting on this interaction, I recognize that while the initial query about congratulating a scholar was broad, the user's follow-up request for a specific graduate sample revealed a need for concrete, emotionally resonant language rather than just general strategies. By providing a structured, warm, and customizable template, I aimed to bridge the gap between abstract advice and practical application, ensuring the user could easily adapt the message to their specific relationship with the graduate. This approach highlights the importance of tailoring responses to the user's evolving needs, moving from high-level concepts to actionable examples that convey genuine pride and support.
\end{examplebox}
\caption{Full dialogue and role-specific reflections for the example in Figure~\ref{fig:reflection-example}. Qwen 3.5 9B generates the two reflections from the user and assistant perspectives using the role-simulation prompt variant.}
\label{fig:reflection-example-full}
\end{figure}

\subsubsection{Reflection Validation}
\label{app:validation-prompt}

Each generated reflection is scored by GPT 5 Mini using the two-message prompt in Appendix~\ref{app:prompt-reflection-validation}.
Only reflections judged to represent the assigned perspective strongly or weakly are retained for activation extraction.

\subsection{Difference-in-Means Computation}

\subsubsection{Activation Extraction and Aggregation}

For each retained reflection, we record the activation at the first response token for every model layer.
User and assistant reflections are paired by dialogue and prompt variant, and the user-minus-assistant differences are averaged across retained pairs.
The resulting direction is normalized before steering, as described in \S\ref{sec:method-dim}.

\section{What Does the User Role Vector Represent?}
\label{app:representation}

This appendix provides supplementary details for the characterization experiments in \S\ref{sec:repr-style}--\S\ref{sec:repr-traits}.

\subsection{Effect on Communication Style}

\subsubsection{Task Goal and Request Generation}
\label{app:stimulus}

The communication-style experiment in \S\ref{sec:repr-style} steers the simulator while it generates a request message conditioned on a task goal.
We generate diverse task goals using the prompt in Appendix~\ref{app:prompt-task-goal} and, for each goal, prompt the steered and unsteered simulators to produce a request message using the prompt in Appendix~\ref{app:prompt-request-generation}.
The prompted baseline uses the user-like request prompt in Appendix~\ref{app:prompt-user-like-request-generation}.

\subsubsection{Steering Parameter Selection}
\label{app:parameter-selection}

We assess the steering range by sweeping layers 10--14 and $\alpha \in \{0.1,0.2,0.3,0.4\}$ on Qwen 3.5 9B, using 100 task goals per condition and averaging five judge evaluations per request.
Above $\alpha=0.3$, the simulator produces extremely short messages that largely repeat the task goal.
Table~\ref{tab:parameter-selection} reports the communication-style score and ROUGE-L precision against the task goal, where higher precision indicates greater reuse of goal text.
Both values rise with steering strength: at layer~11, increasing $\alpha$ from 0.1 to 0.3 raises the judge score from 1.920 to 3.385 but also raises precision from .295 to .625; moving to $\alpha=0.4$ yields a smaller score gain while precision reaches .709.
Reliability also differs across models: at layer~11, Granite 3.3 8B's repeated-token error rate rises from 0\% at $\alpha\leq0.2$ to 13\% at $\alpha=0.3$ and 91\% at $\alpha=0.4$.
We therefore use $\alpha \in \{0.1,0.2,0.3\}$ to span low-to-high intervention levels before this degradation becomes severe.

\begin{table}[!ht]
    \centering
    \footnotesize
    \setlength{\tabcolsep}{1.5pt}
    \begin{tabular*}{\columnwidth}{@{\extracolsep{\fill}}lcccc@{}}
        \toprule
        \textbf{Layer} & $\alpha=0.1$ & $\alpha=0.2$ & $\alpha=0.3$ & $\alpha=0.4$ \\
        \midrule
        10 & 1.780 / .271 & 2.224 / .380 & 3.003 / .540 & 3.547 / .647 \\
        11 & 1.920 / .295 & 2.802 / .494 & 3.385 / .625 & 3.645 / .709 \\
        12 & 1.953 / .323 & 3.126 / .553 & 3.560 / .626 & 3.880 / .705 \\
        13 & 1.959 / .304 & 3.050 / .513 & 3.553 / .578 & 3.827 / .621 \\
        14 & 1.683 / .243 & 2.098 / .292 & 2.695 / .372 & 3.417 / .465 \\
        \bottomrule
    \end{tabular*}
    \caption{Communication-style scores and task-goal overlap across steering parameters (judge score / ROUGE-L precision). Stronger steering improves user-likeness but also increases reuse of task-goal text.}
    \label{tab:parameter-selection}
\end{table}

\subsubsection{Task-Goal Generalization}
\label{app:goal-generalization}

\textbf{Setup.}
We repeat the request-generation experiment with three external user request datasets: MultiWOZ 2.0 and the 2021 and 2022 TREC Health Misinformation topic sets~\cite{budzianowski2018multiwoz,clarke2021trechealth,clarke2022trechealth}.
From the MultiWOZ test split, we randomly sample 100 of 1{,}000 eligible dialogue goals and instruct the simulator to reveal each goal gradually; the recorded first utterance and dialogue are withheld.
For TREC Health, we use all 50 topics from each year and provide only the topic query and background, withholding the gold question.
For each source, Qwen 3.5 9B generates one message at layer~11 under $\alpha \in \{0,0.1,0.2,0.3\}$ with deterministic decoding.
Each generation is scored once by the communication-style judge in \S\ref{app:judge}.
Table~\ref{tab:goal-generalization} reports the resulting scores; the generation prompts appear in \S\ref{app:prompt-goal-generalization-multiwoz}.

\begin{table}[!ht]
\centering
{%
\footnotesize
\setlength{\tabcolsep}{1.25pt}
\begin{tabular*}{\columnwidth}{@{}l@{\extracolsep{\fill}}ccccc@{}}
\toprule
\textbf{Source} & $\boldsymbol{\alpha}$ & \textbf{Overall} & \textbf{Brevity} & \textbf{Informality} & \shortstack{\textbf{Information}\\\textbf{pacing}} \\
\midrule
MultiWOZ & 0.0 & 1.467 & 1.510 & 1.590 & 1.300 \\
         & 0.1 & 1.777 & 1.890 & 1.800 & 1.640 \\
         & 0.2 & 2.077 & 2.120 & 1.940 & 2.170 \\
         & 0.3 & 2.847 & 2.840 & 2.620 & 3.080 \\
\midrule
TREC     & 0.0 & 2.550 & 3.260 & 1.540 & 2.850 \\
         & 0.1 & 2.987 & 3.670 & 1.890 & 3.400 \\
         & 0.2 & 3.090 & 3.730 & 2.050 & 3.490 \\
         & 0.3 & 3.533 & 4.170 & 2.300 & 4.130 \\
\bottomrule
\end{tabular*}
}
\caption{User-likeness scores across steering strengths for MultiWOZ and TREC task goals. Scores increase monotonically with steering strength on both datasets.}
\label{tab:goal-generalization}
\end{table}

\textbf{Results.}
The effect replicates on human-authored goals in MultiWOZ 2.0: the overall score rises monotonically from 1.47 to 2.85, and all three dimensions improve with steering strength.
Information pacing shows the largest change (1.30 to 3.08), indicating that steering helps the simulator avoid revealing the full multi-turn goal in its opening message.

The same monotonic pattern holds for health information needs in TREC, with the overall score increasing from 2.55 to 3.53.
Informality remains the lowest dimension (1.54 to 2.30), while brevity and information pacing rise to 4.17 and 4.13, respectively.
Although the goal domain affects the request content, the direction of the steering effect remains consistent.

\subsubsection{Communication-Style Evaluation}
\label{app:judge}

We use GPT 5 Mini as the judge for scoring user-likeness along three dimensions: brevity, informality, and information pacing.
The judge scores each dimension on a 1--5 scale, with higher scores indicating more user-like style.
The full prompt is shown in Appendix~\ref{app:prompt-user-likeness-judge}.

\subsubsection{LLM-as-Judge Validation}
\label{app:judge-validation}

We validate that the judge separates real user messages from strong LLM-generated approximations of the same message.

\textbf{Setup.}
We sample 1{,}000 English WildChat conversations with 2--10 turns and at most 1{,}000 words per message.
For each conversation, we hide the final user message and prompt GPT 5 to reconstruct the most likely next user message from the prior dialogue.
The real message and the reconstructed message are then scored independently by the judge.
Two samples are excluded due to generation errors.

\begin{figure}[!t]
\centering
\includegraphics[width=\columnwidth]{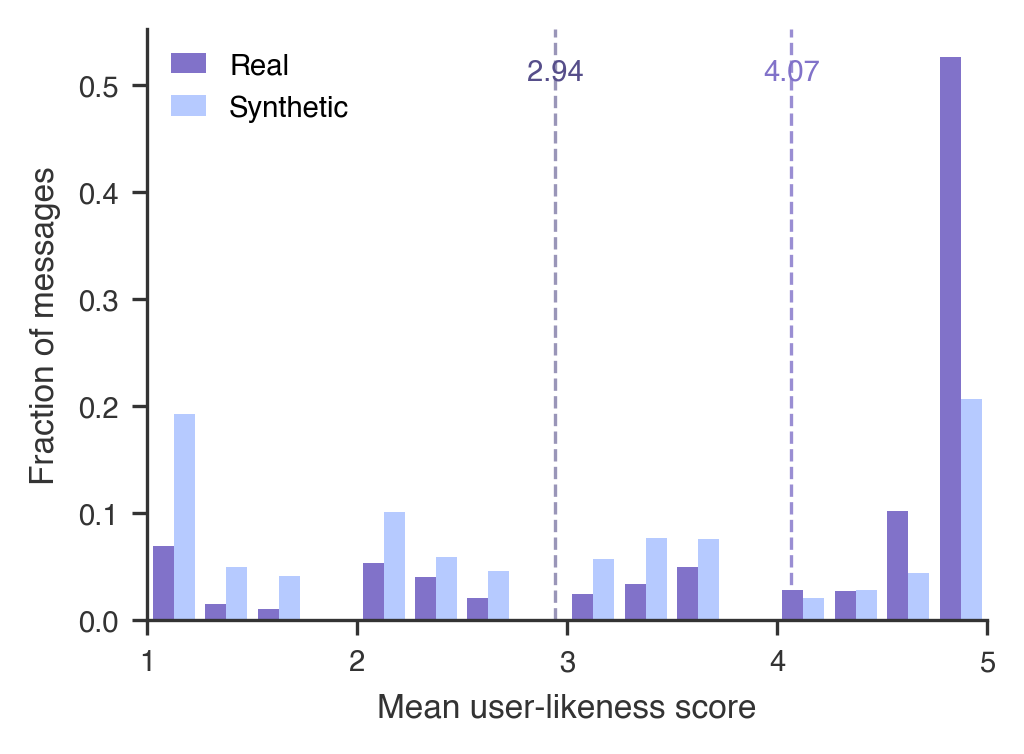}
\caption{Distribution of user-likeness scores for 998 paired real and reconstructed WildChat messages. Real messages score higher; dashed lines mark the group means.}
\label{fig:judge-distribution}
\end{figure}

\begin{figure}[!t]
\centering
\includegraphics[width=\columnwidth]{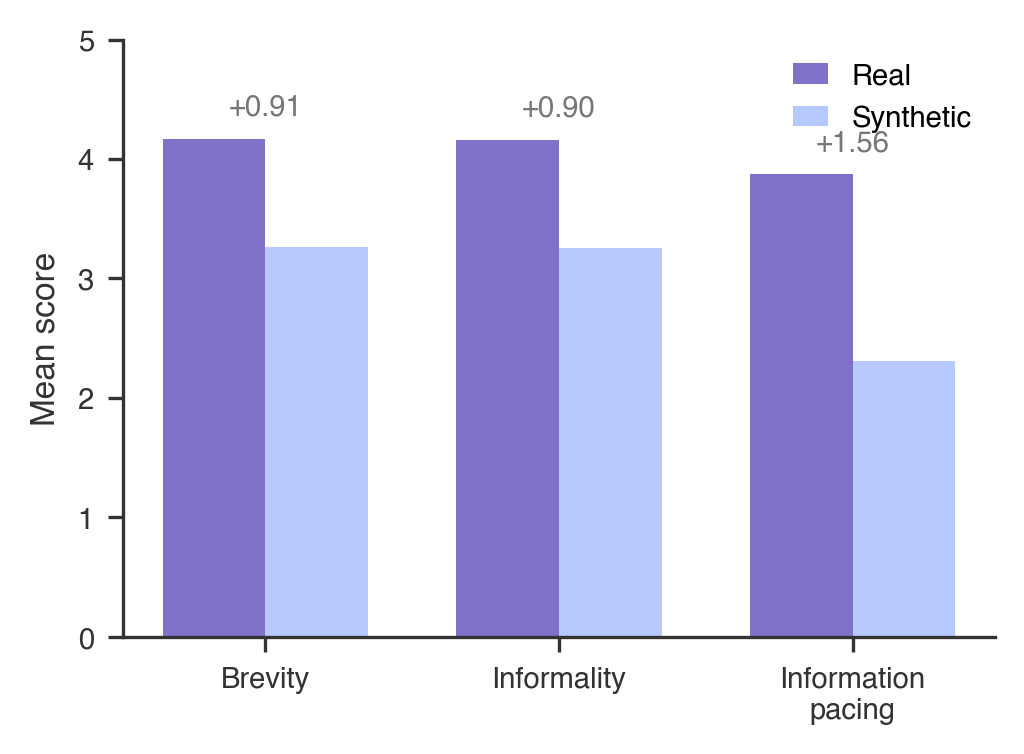}
\caption{User-likeness scores by dimension for 998 paired real and reconstructed WildChat messages. Real messages score higher on all dimensions, with the largest gap in information pacing; labels show real-minus-reconstructed mean differences.}
\label{fig:judge-dimensions}
\end{figure}

\textbf{Results.}
On the remaining 998 paired samples, real user messages receive a mean user-likeness score of 4.07 against 2.94 for synthetic messages, a mean gap of 1.13 (95\% CI: 1.02--1.22) (Figure~\ref{fig:judge-distribution}).
Real messages win the pairwise comparison in 63.2\% of cases, tie in 24.9\%, and lose in 11.8\%.
The gap is consistent across all three dimensions: brevity (+0.91), informality (+0.90), and information pacing (+1.56), with the largest separation on information pacing (Figure~\ref{fig:judge-dimensions}).
The result confirms that the metric distinguishes real user messages from strong LLM-generated approximations of the same message.

\subsubsection{Human Validation}
\label{app:judge-human-validation}

We further test whether the judge agrees with human ratings of generated requests.

\textbf{Setup.}
We sample 200 requests from 50 task goals across four Qwen 3.5 9B conditions at layer~11, with $\alpha \in \{0, 0.1, 0.2, 0.3\}$.
Thirty participants each rate one set of 40 requests on the same 1--5 brevity, informality, and information-pacing scales used by the judge.
The five sets are disjoint, and six participants rate each request, yielding 1{,}200 ratings.
We measure human--judge alignment using Spearman correlation after averaging the six ratings for each request.
Tie-corrected Kendall's $W$ measures agreement among individual human raters.

Participants were recruited through Prolific for a 20-minute study and paid \pounds5.40 each.
Figure~\ref{fig:human-validation-instructions} shows the full instructions and consent notice given to participants.
No personal information were collected, and payments were administered through pseudonymous Prolific participant IDs.

\begin{figure}[!ht]
\centering
\scriptsize
\begin{examplebox}
\textbf{User Simulator Evaluation}\\[2pt]
In this study, you will read 40 request messages generated by a user simulator (a simulator designed to mimic user behavior) and rate how human-like each message feels. There are no right or wrong answers. Please assign scores based on the rubric provided.\\[4pt]
\textbf{Information and consent}\\[2pt]
\textbullet\ Your Prolific participant ID is collected to manage participation and payment.\\
\textbullet\ No name, email address, or other directly identifying information is requested.\\
\textbullet\ Only de-identified ratings may be released for research purposes.\\
\textbullet\ Participation is voluntary. You may stop before submitting the study.\\[3pt]
\texttt{[ ]} I have read the information above and consent to participate.
\end{examplebox}

\begin{examplebox}
\textbf{How to evaluate a message}\\[2pt]
Read the generated request, then select one score for each of the three style dimensions. Use the descriptions and examples shown with each rubric.\\[3pt]
\textit{Example request:} ``best headphones for flights?''\\
\textit{Example ratings:} Brevity: 5 (Short); Informality: 5 (Casual); Information pacing: 5 (Concise).\\
This short, casual request gives only the immediate need, so it receives a high score on all three dimensions.
\end{examplebox}

\begin{examplebox}
\textbf{Brevity.} Focus on word count and short-turn behavior.\\
\textit{1 (Long):} Long, complete, polished, or over-explained. Example: ``Could you provide a detailed and well-structured comparison of three noise-canceling headphones, including comfort, battery life, price, and suitability for travel?''\\
\textit{5 (Short):} Very short, direct, fragment-like, or only says what is needed. Example: ``best headphones for flights?''\\[3pt]

\textbf{Informality.} Focus on surface style, formality, and conversational roughness.\\
\textit{1 (Formal):} Formal, careful, service-like, or written in polished complete sentences. Example: ``Could you please assist me in drafting an appropriate email declining this invitation?''\\
\textit{5 (Casual):} Casual, blunt, uneven, ordinary, or written like quick everyday conversation. Example: ``how do i say no to this invite without sounding rude''\\[3pt]

\textbf{Information pacing.} Focus on whether the message front-loads information or withholds details.\\
\textit{1 (Dense):} Gives the full context, identifiers, constraints, and desired outcome upfront. Example: ``I need to return order \#48291, a black jacket bought last Tuesday, because the zipper broke; please write a message asking for a refund or replacement.''\\
\textit{5 (Concise):} Gives only the immediate need, leaves details implicit, or would require follow-up. Example: ``can i return this?''
\end{examplebox}
\caption{Instructions and consent notice shown to participants in the human-validation study. Participants had to select the consent checkbox before beginning.}
\label{fig:human-validation-instructions}
\end{figure}

\textbf{Results.}
The mean human score rises monotonically from 2.23 without steering to 3.77 at $\alpha=0.3$, matching the judge's increase from 1.63 to 3.39 (Table~\ref{tab:human-judge-validation}).
Human and judge scores correlate strongly overall ($\rho=0.800$), with dimension-level correlations of 0.798 for brevity, 0.439 for informality, and 0.754 for information pacing.
Human raters also show substantial agreement overall ($W=0.700$), with $W=0.800$ for brevity, 0.432 for informality, and 0.579 for information pacing.

\begin{table}[!t]
    \centering
    \footnotesize
    \setlength{\tabcolsep}{1.5pt}
    \begin{tabular*}{\columnwidth}{@{\extracolsep{\fill}}clcccc@{}}
        \toprule
        $\boldsymbol{\alpha}$ & \textbf{Rater} & \textbf{Brevity} & \textbf{Informality} & \shortstack{\textbf{Information}\\\textbf{pacing}} & \textbf{Overall} \\
        \midrule
        0.0 & Human     & 2.29 & 2.18 & 2.23 & 2.23 \\
            & LLM judge & 1.80 & 1.02 & 2.08 & 1.63 \\
        \addlinespace[1pt]
        0.1 & Human     & 2.72 & 2.60 & 2.54 & 2.62 \\
            & LLM judge & 2.26 & 1.20 & 2.20 & 1.89 \\
        \addlinespace[1pt]
        0.2 & Human     & 3.51 & 3.00 & 3.36 & 3.29 \\
            & LLM judge & 3.28 & 1.28 & 3.72 & 2.76 \\
        \addlinespace[1pt]
        0.3 & Human     & 4.15 & 3.32 & 3.83 & 3.77 \\
            & LLM judge & 4.18 & 1.50 & 4.50 & 3.39 \\
        \bottomrule
    \end{tabular*}
    \caption{Human and LLM-judge communication-style scores across steering strengths. Both assign higher scores as steering increases.}
    \label{tab:human-judge-validation}
\end{table}

\subsubsection{Activation Correlation with Communication Style}
\label{app:communication-style-activation-correlation}

\textbf{Setup.}
We use the same 100 task goals as the communication-style experiment in \S\ref{sec:repr-style} and generate one unsteered request per goal with Qwen 3.5 9B.
The same GPT 5 Mini judge scores each request for brevity, informality, and information pacing (\S\ref{app:judge}).
We teacher-force each request without steering, project its layer-11 hidden states onto the normalized user-minus-assistant direction, and average the projections across response tokens.

\textbf{Results.}
Mean user-role activation is positively correlated with the overall communication-style score (Pearson's $r=.260$; Spearman's $\rho=.313$; Table~\ref{tab:communication-style-activation-correlation}).
The rank association is strongest for brevity ($\rho=.322$).
To account for response length, we residualize the activation and style scores against log token count for Pearson correlation, and residualize their ranks against ranked token count for Spearman correlation.
After this control, the overall association becomes slightly stronger ($r=.292$; $\rho=.337$); in this setup, response length is associated with communication-style scores but only weakly with user-role activation.

\begin{table}[!ht]
    \centering
    \footnotesize
    \begin{tabular*}{\columnwidth}{@{\extracolsep{\fill}}lrrrr@{}}
        \toprule
        & \multicolumn{2}{c}{\textbf{Correlation}} & \multicolumn{2}{c}{\textbf{Length-controlled}} \\
        \cmidrule(lr){2-3}\cmidrule(l){4-5}
        \textbf{Metric} & \textbf{$r$} & \textbf{$\rho$} & \textbf{$r$} & \textbf{$\rho$} \\
        \midrule
        Overall            & .260 & .313 & .292 & .337 \\
        Brevity            & .208 & .322 & .247 & .379 \\
        Informality        & .224 & .241 & .223 & .233 \\
        Information pacing & .219 & .244 & .219 & .239 \\
        \bottomrule
    \end{tabular*}
    \caption{Correlation between user-role activation and communication-style scores. Length-controlled columns report partial correlations; all associations remain positive.}
    \label{tab:communication-style-activation-correlation}
\end{table}

\subsubsection{Layer-Wise Steering Effects}
\label{app:layer-effect}

We sweep the injection layer~$\ell \in [1,31]$ and the steering coefficient $\alpha \in \{0.1, 0.2, 0.3\}$, with 100 sampled requests per (layer, $\alpha$) cell scored by the LLM-as-judge in \S\ref{app:judge}.
The unsteered baseline (alpha-zero) is layer-independent because $\alpha=0$ zeros the injected vector; we report the empirical baseline, $\bar{s}=1.51$.

\begin{figure}[!t]
\centering
\includegraphics[width=\columnwidth]{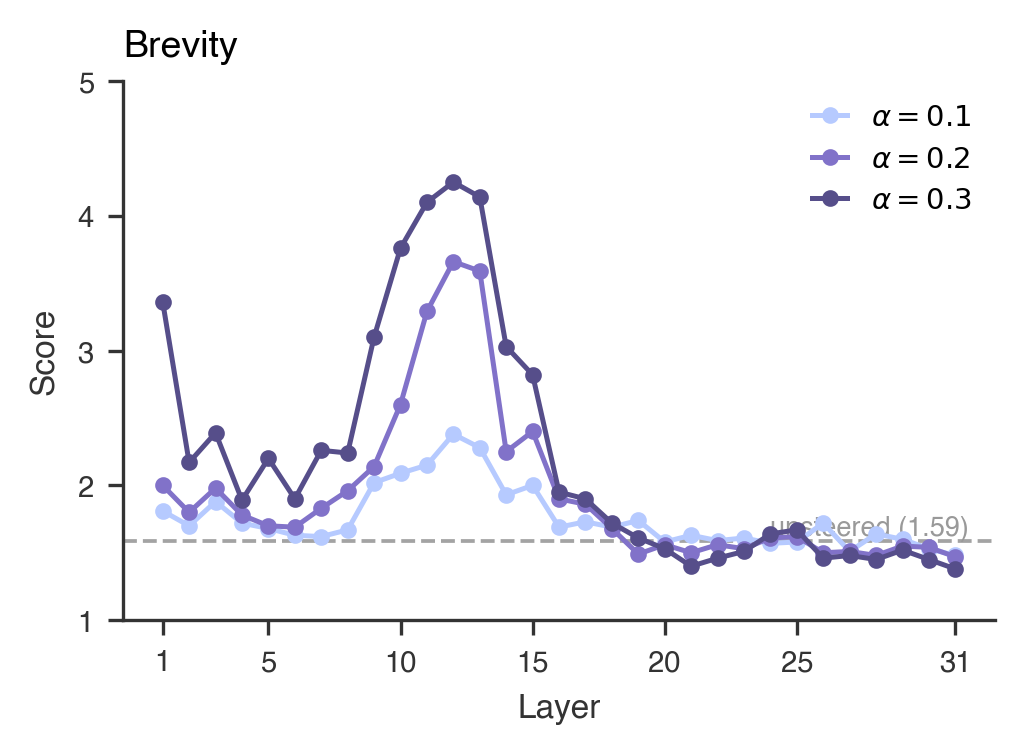}
\caption{Brevity scores across steering layers and strengths. Scores peak at layers 11--13.}
\label{fig:layer-effect-dim-brevity}
\end{figure}

\begin{figure}[!t]
\centering
\includegraphics[width=\columnwidth]{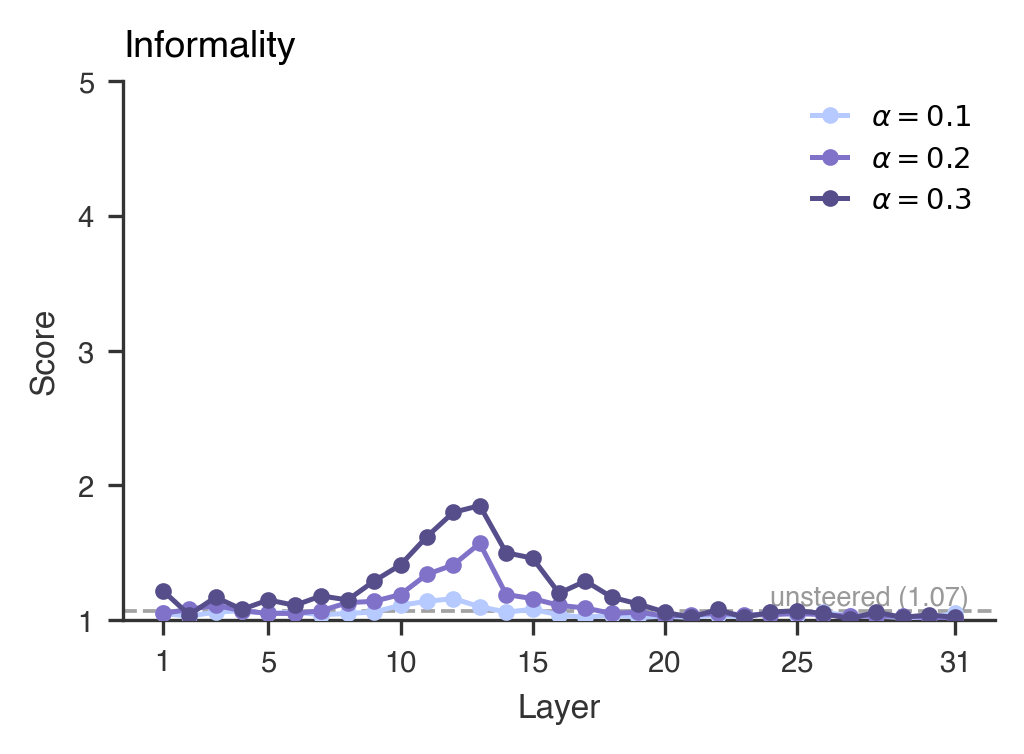}
\caption{Informality scores across steering layers and strengths. Scores rise sharply at layers 11--13 and saturate near the top of the scale.}
\label{fig:layer-effect-dim-informality}
\end{figure}

\begin{figure}[!t]
\centering
\includegraphics[width=\columnwidth]{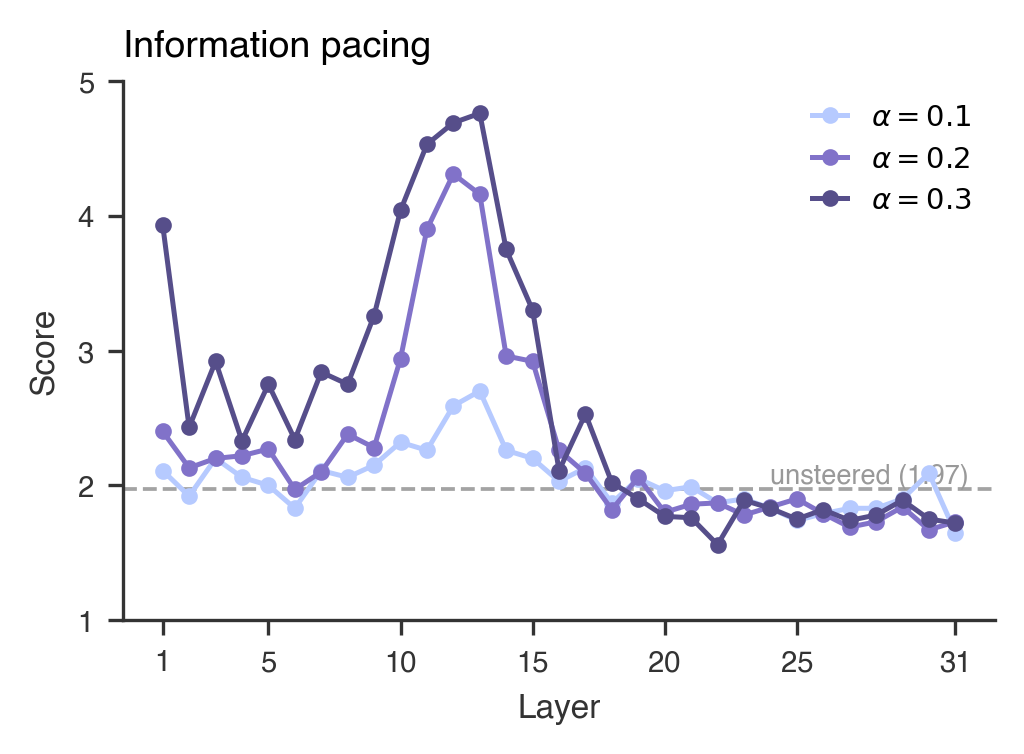}
\caption{Information-pacing scores across steering layers and strengths. This dimension shows the largest shift, peaking near 4.7 at $\alpha=0.3$ in layers 11--13.}
\label{fig:layer-effect-dim-information-pacing}
\end{figure}

\textbf{Per-dimension breakdown.}
Figures~\ref{fig:layer-effect-dim-brevity}--\ref{fig:layer-effect-dim-information-pacing} report the per-dimension scores across layers.
All three dimensions respond strongly in the layer 11--13 band.
Information pacing has the largest absolute shift (peaking near 4.7), informality reaches the top of the scale earliest and saturates, and brevity tracks the aggregate score most closely.
Layers 19 and beyond receive little or no benefit from stronger~$\alpha$ on any dimension.

\textbf{Steering coefficient.}
Within the layer 11--13 peak, $\alpha=0.3$ outperforms smaller coefficients without showing saturation: the gap between $\alpha=0.2$ and $\alpha=0.3$ is 0.45--0.57 points at layers 11, 12, and 13.
Outside this band, the response is brittle: at layer 16 and later, $\alpha=0.3$ often scores no better than $\alpha=0.1$, and at layers 19--31 it is sometimes worse.
We treat layer and~$\alpha$ as a coupled choice rather than independent hyperparameters: stronger~$\alpha$ is only useful within the mid-layer window.

\subsubsection{Cross-Model Steering Effects}
\label{app:layer-effect-across-models}

\begin{figure}[!t]
\centering
\includegraphics[width=\columnwidth]{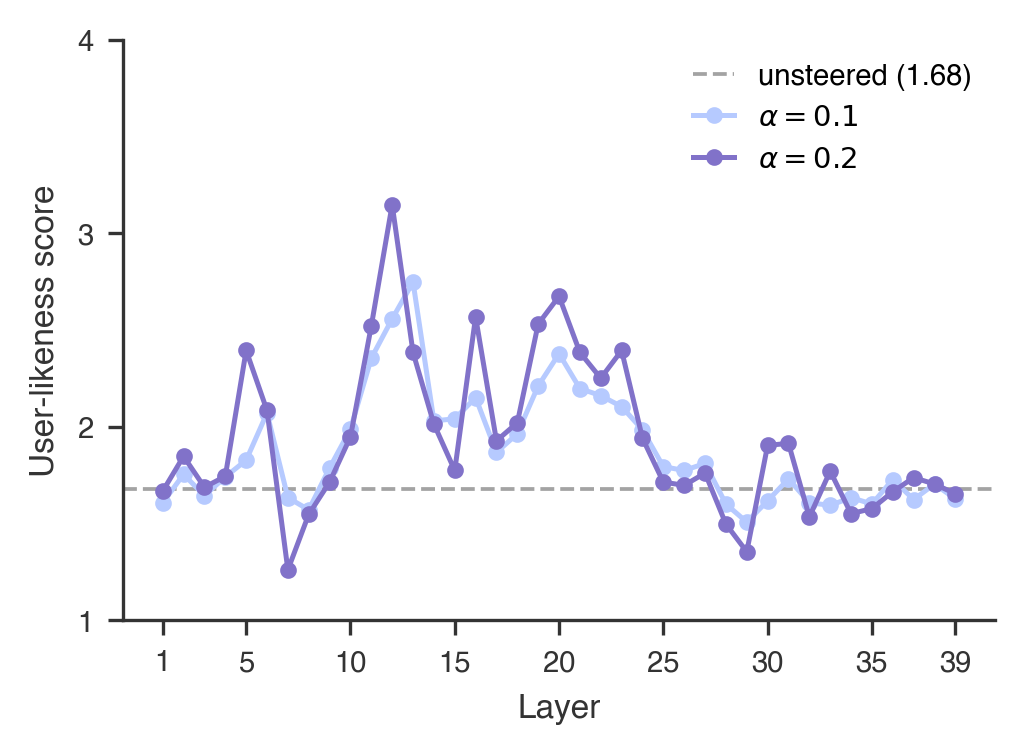}
\caption{User-likeness scores across steering layers and strengths for Granite 3.3 8B. Effects are strongest near layer~12 and layers 19--23. The $\alpha=0.3$ condition is omitted because it produces degenerate outputs.}
\label{fig:layer-effect-granite}
\end{figure}

\begin{figure}[!t]
\centering
\includegraphics[width=\columnwidth]{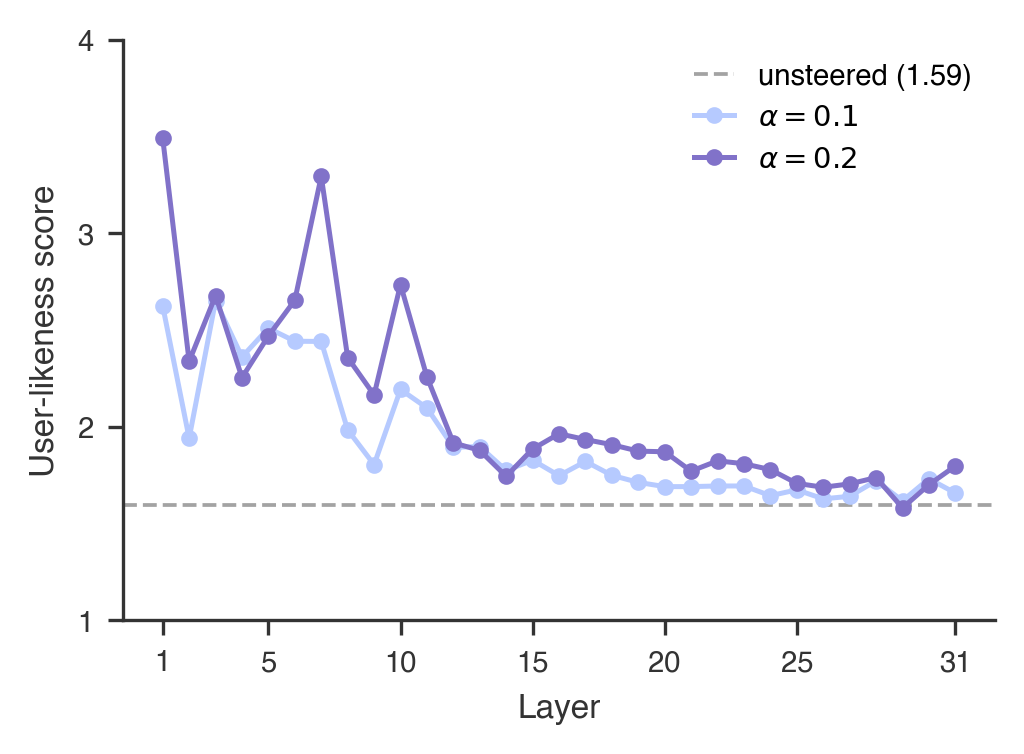}
\caption{User-likeness scores across steering layers and strengths for Llama 3.1 8B. Effects are strongest in layers 1--10 and decay after layer~12. The $\alpha=0.3$ condition is omitted because it produces degenerate outputs.}
\label{fig:layer-effect-llama}
\end{figure}

We repeat the layer sweep on Granite 3.3 8B and Llama 3.1 8B using the same request-generation setting, with first-token role vectors applied to all generated tokens.
Both models show that user-role steering has its largest effect in early-to-mid layers, although the exact peak differs by architecture.
For Granite, the more stable gains appear around layer~12 and layers 19--23 (Figure~\ref{fig:layer-effect-granite}); for Llama, the strongest gains appear earlier, mostly in layers 1--10 (Figure~\ref{fig:layer-effect-llama}).
This supports the main result that user-like message style is most sensitive to earlier parts of the model rather than late layers.

At the same time, reliability differs across models.
At $\alpha=0.3$, both Granite and Llama sometimes produce degenerate outputs, including repeated words and meaningless tokens.
We therefore exclude these unstable settings from the summary table and treat the cross-model result as qualitative evidence: the layer-localized effect is coherent across models, but the reliable steering strength is model-dependent.

\subsubsection{Layer Structure of the User Role Direction}
\label{app:layer-structure}

\begin{figure}[!ht]
\centering
\includegraphics[width=\columnwidth]{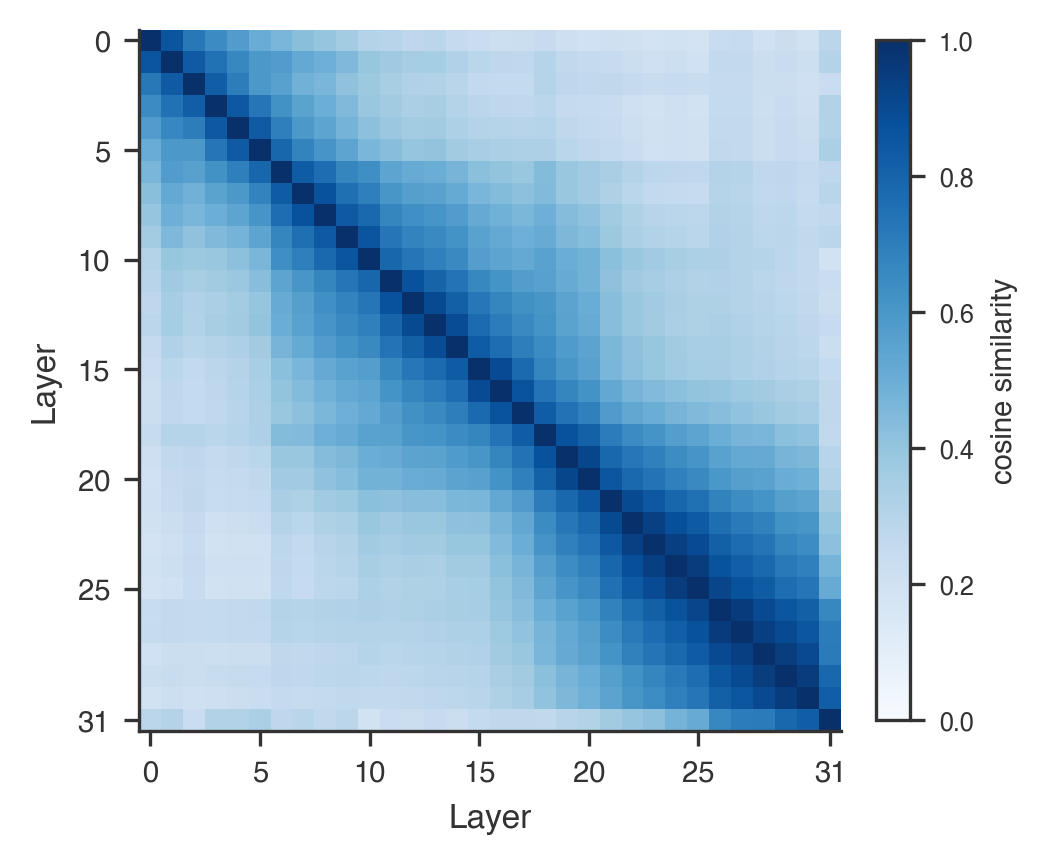}
\caption{Layer-to-layer cosine similarity of the user-role direction in Qwen 3.5 9B. Layers 6--20 form a broad mid-layer cluster.}
\label{fig:layer-structure-qwen}
\end{figure}

The layer-wise steering effect is not simply a property of one isolated injection layer. 
In Qwen 3.5 9B, the user role direction forms local bands in the layer-to-layer cosine matrix (Figure~\ref{fig:layer-structure-qwen}). 
Adjacent layers are generally similar, and layers 6--20 form a broad mid-layer cluster, with a slight concentration around layers 11--13. This structure helps explain why the main effect in \S\ref{sec:repr-style} appears around this broader part of the network rather than at a single isolated layer. 
Steering in this range appears to act on a user role direction that is well matched to the layer-local representation, rather than on a direction that is equally effective across all layers.

\begin{figure}[!t]
\centering
\includegraphics[width=\columnwidth]{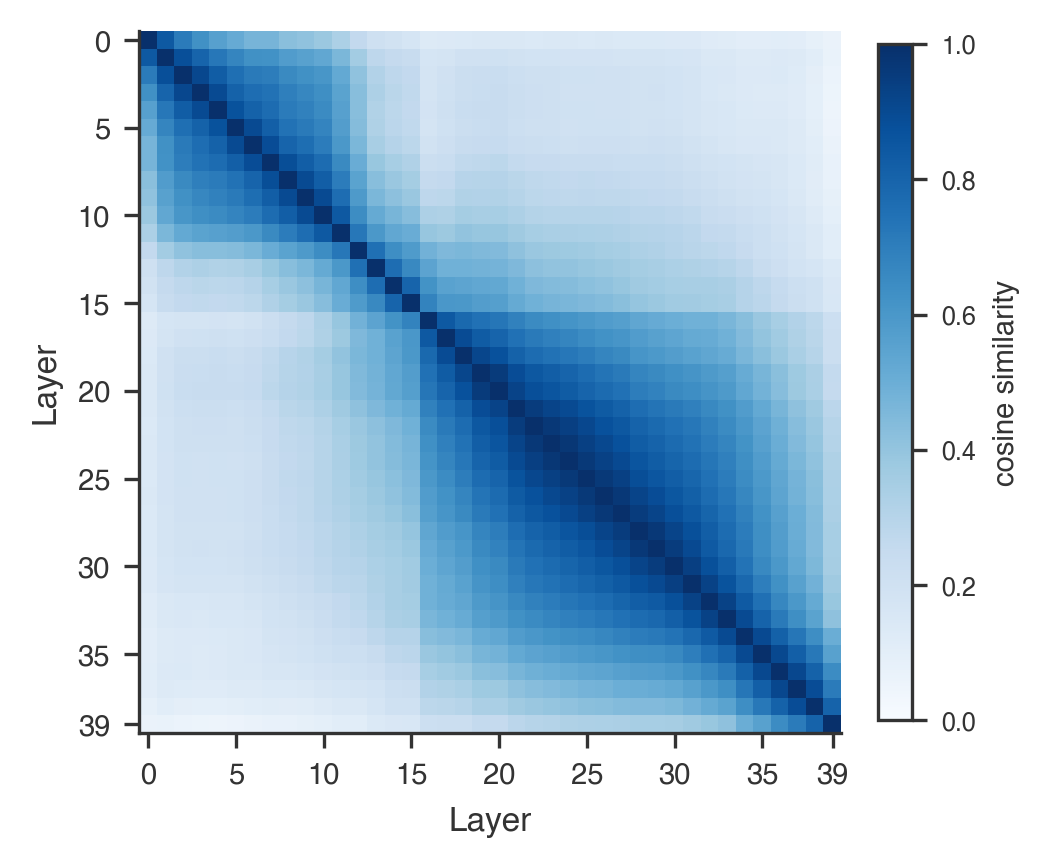}
\caption{Layer-to-layer cosine similarity of the user-role direction in Granite 3.3 8B. The matrix separates into early and late blocks, with a transition around layers 12--16.}
\label{fig:layer-structure-granite}
\end{figure}

\begin{figure}[!t]
\centering
\includegraphics[width=\columnwidth]{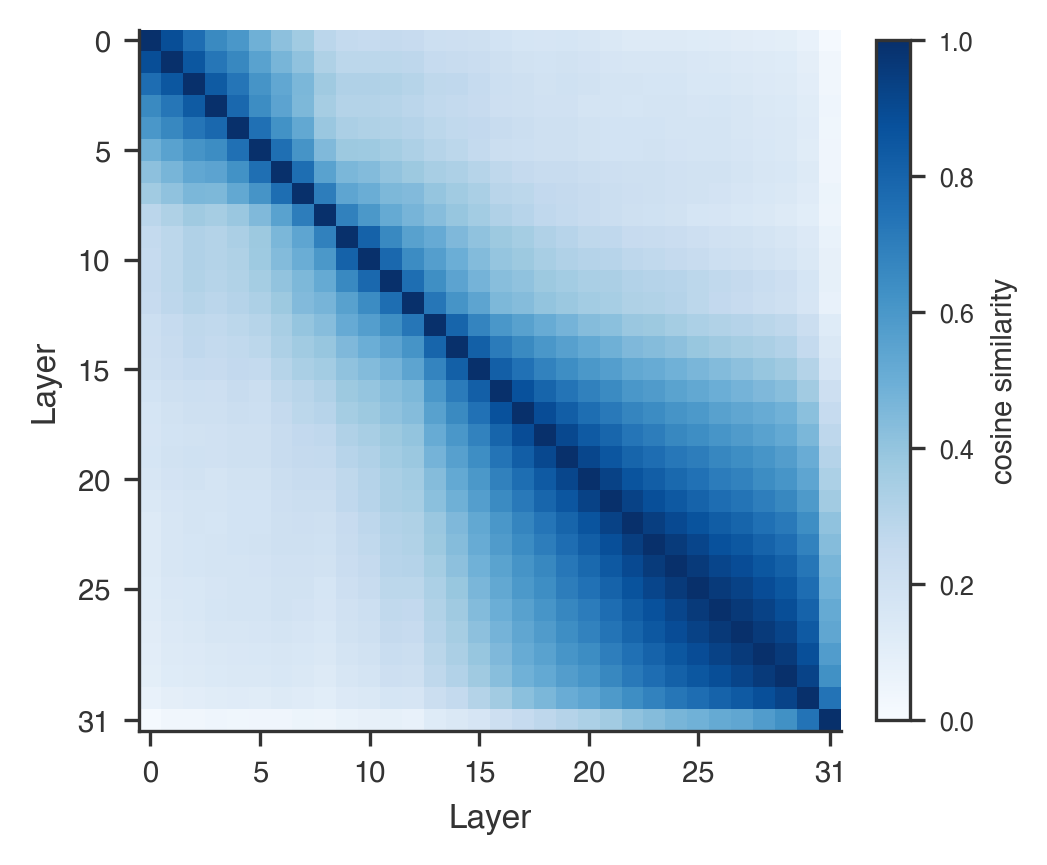}
\caption{Layer-to-layer cosine similarity of the user-role direction in Llama 3.1 8B. Early layers are fragmented, while later layers form a more stable block.}
\label{fig:layer-structure-llama}
\end{figure}

The cross-model matrices show the same principle with slightly different boundaries. 
Granite 3.3 8B separates into a broad early block and a later block, with a transition around layers 12--16; its strongest steering effects occur near this transition and again around layers 19--23 (Figure~\ref{fig:layer-structure-granite}). 
Llama 3.1 8B shows a more fragmented early structure, with weaker continuity across layers 8--13 and a more stable later block (Figure~\ref{fig:layer-structure-llama}). 
Together with the layer-sweep results in \S\ref{app:layer-effect-across-models}, these matrices suggest that user role steering is effective on specific layer-local representation of the user role. 
The relevant layer range is therefore conceptually coherent, where user-like message style is more sensitive to earlier or transitional parts of the network than to later layers.

\subsubsection{Bidirectional Steering}
\label{app:directionality-sweep}

The directionality control in \S\ref{sec:repr-style} is a coarse comparison between user-direction and assistant-direction steering at $\alpha=0.3$.
To check that the effect is smooth and monotonic, we sweep the signed steering coefficient $\alpha \in \{-0.3, -0.2, -0.1, 0.0, 0.1, 0.2, 0.3\}$ at layer~11, where positive $\alpha$ steers along the user direction and negative $\alpha$ steers along the assistant direction.
Each cell uses the same 100 task goals and the same judge as the main experiment.

\begin{figure}[!t]
\centering
\includegraphics[width=0.9\columnwidth]{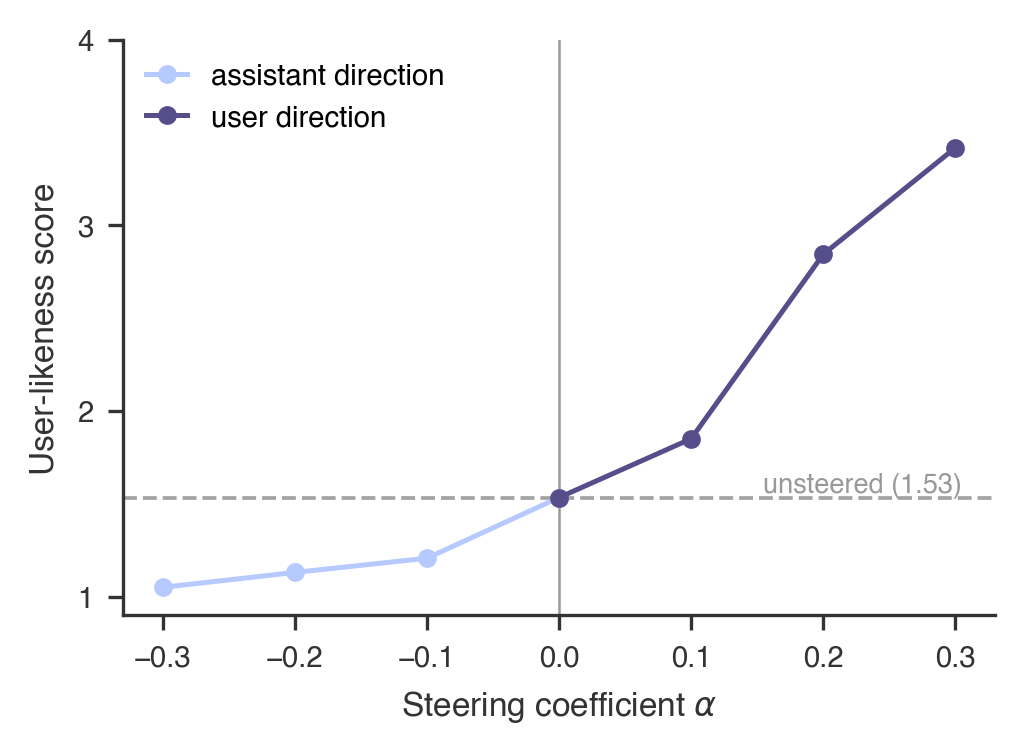}
\caption{User-likeness scores under bidirectional steering at layer~11. Scores increase monotonically from assistant-direction (negative $\alpha$) to user-direction (positive $\alpha$) steering.}
\label{fig:directionality-sweep}
\end{figure}

The user-likeness score increases monotonically with $\alpha$ across the full range (Figure~\ref{fig:directionality-sweep}).
The unsteered baseline sits at $1.53$.
Moving toward the user direction lifts the score to $1.85$ at $\alpha=0.1$, $2.84$ at $\alpha=0.2$, and $3.42$ at $\alpha=0.3$.
Moving toward the assistant direction drops the score to $1.21$, $1.13$, and $1.05$ at $\alpha=-0.1$, $-0.2$, and $-0.3$ respectively, approaching the floor of the 1--5 scale.
The same vector therefore controls user-likeness in both directions, with no inflection or saturation visible on the user side within the swept range.

\subsection{Effect on Interaction Behavior}

\subsubsection{Disengagement Taxonomy}
\label{app:disengagement-taxonomy}

The interaction-behavior experiment in \S\ref{sec:repr-error} samples dialogues stratified by a disengagement taxonomy derived from real user-assistant interaction logs.

\textbf{Source data.}
Publicly available user-assistant interaction logs are collected from GitHub through the GitHub Search API.
The sources include coding-agent frameworks such as SpecStory, Aider, and Claude Code.
Of the 33{,}033 collected logs, 9{,}465 are excluded for empty sessions, incomplete conversations (e.g., a single message), or corrupted files.
Logs with 2 to 50 turns are retained, yielding 16{,}077 interaction logs for the taxonomy analysis.

\textbf{Taxonomy construction.}
Gemini 2.5 Flash develops an initial codebook of user disengagement triggers by iterating over each dialogue.
For each dialogue, the model either identifies a new disengagement reason or merges it into an existing code.
The first pass produces 61 unique disengagement triggers.
Two researchers then revisit the codebook and merge codes by definition and sampled examples, producing eight concrete disengagement types (Table~\ref{tab:disengagement-taxonomy}).
A ninth \emph{Unknown} category captures end-of-conversation events that cannot be attributed to model behavior; it is retained in the taxonomy for completeness but excluded from sampling in \S\ref{sec:repr-error}.

\begin{table}[!t]
\centering
\footnotesize
\renewcommand{\arraystretch}{1.08}
\begin{tabular}{@{}p{\columnwidth}@{}}
\toprule
\textbf{Normal completion} \\
The user's goal is fully addressed and the conversation ends naturally, with no further interaction required. \\
\addlinespace[0.25em]
\textbf{Forgetting} \\
The model fails to incorporate prior context or instructions, losing track of constraints, earlier instructions, or dialogue history. \\
\addlinespace[0.25em]
\textbf{Technical failure} \\
The model fails to produce a correct or functional output, such as code that errors, an incomplete generation, or an execution failure. \\
\addlinespace[0.25em]
\textbf{Misinterpretation} \\
The model misunderstands the user's intent, requirements, or phrasing, yielding a response that is off-target or irrelevant. \\
\addlinespace[0.25em]
\textbf{Expectation mismatch} \\
The model's output is technically plausible but diverges from the user's assumed workflow, conventions, or mental model of the task. \\
\addlinespace[0.25em]
\textbf{Unsolicited action} \\
The model performs actions beyond what is explicitly requested, introducing edits, additions, or suggestions outside the user's scope. \\
\addlinespace[0.25em]
\textbf{Refusal} \\
The model declines or deflects a reasonable request, defers unnecessarily to the user, or adds friction through excessive clarification or caveats. \\
\addlinespace[0.25em]
\textbf{Hallucination} \\
The model produces factually incorrect or fabricated content, including false claims, fabricated references, or unwarranted confidence. \\
\midrule
\textbf{Unknown} \\
The conversation ends for reasons unrelated to model behavior, such as user-side factors or external circumstances. Excluded from sampling. \\
\bottomrule
\end{tabular}
\caption{Taxonomy of user disengagement triggers derived from public interaction logs. The first eight categories are used for stratified sampling; \emph{Unknown} is excluded because it is not attributable to model behavior.}
\label{tab:disengagement-taxonomy}
\end{table}

\subsubsection{Disengagement Prediction}
\label{app:disengagement-prediction}

The disengagement experiment evaluates whether a simulator can predict when the original user would stop a conversation.
Each dialogue has eight user--assistant turns and an annotated disengagement type from the taxonomy above.
At each prefix turn, the model sees the dialogue up to the current assistant response and chooses whether the user would continue or disengage.
We run two decisions per sample with the option order swapped, so each completed 735-dialogue run contributes 1{,}470 balanced decisions.
The predicted disengagement turn is the first turn where the model chooses disengagement; otherwise, the user is treated as continuing through the final turn.

All conditions use a prompt without a user profile or a request for reasoning (Appendix~\ref{app:prompt-disengagement-prediction}).
For Qwen 3.5 9B, the user-role vector is injected at layer~11 with $\alpha \in \{0,0.1,0.2,0.3\}$ and applied to all generated tokens.
GPT 5.4 and Gemini 3 Flash are evaluated with the same decision prompt but without activation steering.
Exact match measures whether the predicted turn equals the annotated disengagement turn.
Average distance is the absolute turn error and excludes final-continue trials.
Table~\ref{tab:disengagement-prediction-ending-type} reports the same predictions grouped by normal completion versus disrupted endings.

\begin{table}[!t]
\centering
{%
\footnotesize
\setlength{\tabcolsep}{3pt}
\begin{tabular*}{\columnwidth}{@{}p{0.54\columnwidth}@{\extracolsep{\fill}}cc@{}}
\toprule
\textbf{Condition} & \shortstack{\textbf{Normal}\\\textbf{completion}} & \shortstack{\textbf{Disrupted}\\\textbf{ending}} \\
\midrule
Qwen 3.5 9B (Unsteered) & 27.0\% & 51.8\% \\
+ Steered ($\alpha=0.1$) & 34.0\% & 55.0\% \\
+ Steered ($\alpha=0.2$) & 41.0\% & 58.2\% \\
+ Steered ($\alpha=0.3$) & 56.0\% & 68.5\% \\
\midrule
GPT 5.4 & 23.0\% & 26.5\% \\
Gemini 3 Flash & 64.5\% & 58.6\% \\
\bottomrule
\end{tabular*}
}
\caption{Predicted disengagement rates for normal and disrupted endings. Stronger steering increases both rates; disrupted endings aggregate the seven model-failure categories in Table~\ref{tab:disengagement-taxonomy}.}
\label{tab:disengagement-prediction-ending-type}
\end{table}

\subsubsection{Cross-Model Disengagement Prediction}
\label{app:disengagement-prediction-llama}

We repeat the disengagement-prediction experiment with Llama 3.1 8B at layer~7, using the same 735 dialogues and two option-order decisions per dialogue.
The disengagement rate rises monotonically from 15.3\% without steering to 52.7\% at $\alpha=0.3$, reproducing the Qwen trend (Table~\ref{tab:disengagement-prediction-llama}).
Exact match instead peaks at $\alpha=0.1$ (2.6\%), while average distance increases overall from 3.86 to 5.81 turns.
As with Qwen 3.5 9B, stronger steering makes Llama more likely to disengage but does not reliably improve when it disengages.

\begin{table}[!ht]
\centering
{%
\footnotesize
\setlength{\tabcolsep}{3pt}
\begin{tabular*}{\columnwidth}{@{}c@{\extracolsep{\fill}}ccc@{}}
\toprule
$\boldsymbol{\alpha}$ & \shortstack{\textbf{Exact}\\\textbf{match}} & \shortstack{\textbf{Average}\\\textbf{distance}} & \shortstack{\textbf{Disengagement}\\\textbf{rate}} \\
\midrule
0.0 & 23/1{,}470 (1.6\%) & 3.858 & 15.3\% \\
0.1 & 38/1{,}470 (2.6\%) & 4.366 & 33.1\% \\
0.2 & 17/1{,}470 (1.2\%) & 5.813 & 45.9\% \\
0.3 & 27/1{,}470 (1.8\%) & 5.811 & 52.7\% \\
\bottomrule
\end{tabular*}
}
\caption{Llama 3.1 8B disengagement prediction across steering strengths at layer~7. Steering increases the disengagement rate but not timing accuracy.}
\label{tab:disengagement-prediction-llama}
\end{table}

\subsection{Relationship to the Assistant Bias}

\subsubsection{Assistant-Trait Direction Extraction}
\label{app:trait-analysis}

The analysis in \S\ref{sec:repr-traits} compares the user role vector against a set of assistant-trait directions.
Table~\ref{tab:trait-definitions} lists the 20 traits used in this analysis.

\begin{table}[!t]
\centering
\scriptsize
\renewcommand{\arraystretch}{1.02}
\begin{tabular}{@{}p{\columnwidth}@{}}
\toprule
\textbf{cooperative} \\
Willing to work with others, coordinate actions, and support shared progress. \\
\addlinespace[0.12em]
\textbf{helpful} \\
Inclined to provide assistance, useful information, or support when needed. \\
\addlinespace[0.12em]
\textbf{compliant} \\
Tending to accept requests, rules, instructions, or external direction with little resistance. \\
\addlinespace[0.12em]
\textbf{sycophantic} \\
Excessively agreeable, approving, or flattering toward others, especially authority figures. \\
\addlinespace[0.12em]
\textbf{polite} \\
Courteous, respectful, and socially considerate in communication and behavior. \\
\addlinespace[0.12em]
\textbf{patient} \\
Able to tolerate delays, difficulties, repetition, or frustration without becoming visibly upset. \\
\addlinespace[0.12em]
\textbf{structured} \\
Organized, orderly, and systematic in presenting thoughts, actions, or information. \\
\addlinespace[0.12em]
\textbf{explicit} \\
Clear, direct, and unambiguous in expressing intentions, needs, preferences, or information. \\
\addlinespace[0.12em]
\textbf{complete} \\
Thorough and comprehensive, with few missing details or unresolved gaps. \\
\addlinespace[0.12em]
\textbf{verbose} \\
Using more words, details, or explanation than may be necessary. \\
\addlinespace[0.12em]
\textbf{fluent} \\
Smooth, polished, and natural in expression, with few disruptions, errors, or hesitations. \\
\addlinespace[0.12em]
\textbf{generic} \\
Lacking distinctive personal style, specificity, or unusual characteristics. \\
\addlinespace[0.12em]
\textbf{neutral} \\
Emotionally even, restrained, or impartial, without strong affective expression. \\
\addlinespace[0.12em]
\textbf{consistent} \\
Stable and reliable across situations, with little contradiction or unexpected variation. \\
\addlinespace[0.12em]
\textbf{rational} \\
Logical, deliberate, and reason-based in judgment, decision-making, or behavior. \\
\addlinespace[0.12em]
\textbf{goal\_aligned} \\
Focused on pursuing stated aims and avoiding distractions or irrelevant deviations. \\
\addlinespace[0.12em]
\textbf{accommodating} \\
Willing to adjust, compromise, or adapt to others' needs, constraints, or preferences. \\
\addlinespace[0.12em]
\textbf{deferential} \\
Respectful or submissive toward others' authority, expertise, or judgment. \\
\addlinespace[0.12em]
\textbf{explanatory} \\
Inclined to clarify, justify, or elaborate on reasoning, motives, or context. \\
\addlinespace[0.12em]
\textbf{selfless} \\
Prioritizing others' needs, goals, or success over one's own convenience or interests. \\
\bottomrule
\end{tabular}
\caption{Definitions of the 20 assistant-like traits used to extract trait directions.}
\label{tab:trait-definitions}
\end{table}

\textbf{Data generation.}
For each trait, Gemini 3 Flash generates five positive/negative instruction pairs and 40 neutral questions.
The positive instruction asks the model to exhibit the target trait, while the negative instruction asks the model to avoid it and express the opposing behavior.
The questions are written to elicit the trait without naming it directly.

\textbf{Response generation and scoring.}
Qwen 3.5 9B generates two sampled responses for each instruction--question pair.
With five instruction pairs, two polarities, 40 questions, and two repeats, this yields 800 candidate utterances per trait and 16{,}000 candidate utterances across the 20 traits.
GPT 5 Mini then scores each response from 0 to 100 for how strongly it expresses the target trait, using a trait-specific evaluation prompt generated with the data.
The scoring prompt focuses only on the target trait and treats refusals separately.

\textbf{Vector extraction.}
For each trait, the pipeline selects positive and negative responses with a score gap of at least 50 points, requiring at least 10 selected pairs per trait.
In the final run, this filter retains 6{,}370 contrast pairs.
At layer~11, it extracts hidden states from Qwen 3.5 9B and computes the trait direction as the difference between the mean positive activation and the mean negative activation.
The resulting direction is compared with the user role vector using cosine similarity.

\section{What Does the User Role Vector Reveal About Simulation?}
\label{app:simulation-realism-analyses}

\subsection{Steering Effects on Simulation Realism}
\label{app:simulatorarena-details}

\subsubsection{SimulatorArena Evaluation Setup}
\label{app:simulatorarena-prompts}

SimulatorArena uses separate judge prompts for similarity evaluation.
Appendix~\ref{app:prompt-simulatorarena-interaction-eval} show the interaction-style and writing-style evaluation prompts.
Table~\ref{tab:simulatorarena-features} lists the feature dimensions supplied to the evaluator.

\begin{table}[!t]
\centering
\scriptsize
\renewcommand{\arraystretch}{1.04}
\begin{tabular}{@{}p{\columnwidth}@{}}
\toprule
\textbf{Interaction style} \\
\midrule
\textbf{Answer thoroughness} \\
When responding to questions, does the user tend to give detailed, complete responses or brief, minimal answers? \\
\addlinespace[0.14em]
\textbf{Clarification seeking} \\
How frequently does the user ask for examples or additional explanations? \\
\addlinespace[0.14em]
\textbf{Feedback style} \\
Does the user provide ongoing verbal feedback (\textit{okay}, \textit{I see}, etc.)? \\
\addlinespace[0.14em]
\textbf{Topic coherence} \\
How well does the user maintain focus on a single topic versus jumping between different points? \\
\addlinespace[0.14em]
\textbf{Contextual reference} \\
Does the user refer back to previous messages or maintain conversation context? \\
\addlinespace[0.14em]
\textbf{Implementation of feedback} \\
How does the user respond to and implement suggestions or corrections? \\
\addlinespace[0.14em]
\textbf{Confidence level} \\
How does the user express their confidence or uncertainty in their understanding? \\
\addlinespace[0.14em]
\textbf{Emotional expression} \\
How does the user communicate their emotional state, such as frustration or enthusiasm? \\
\addlinespace[0.14em]
\textbf{Real-time thought expression} \\
How does the user articulate their thought process in real time, reflecting their immediate understanding and confusion? \\
\addlinespace[0.14em]
\textbf{Error handling} \\
How does the user demonstrate trial and error, acknowledge mistakes, and correct them? \\
\addlinespace[0.14em]
\textbf{Structured problem solving} \\
Does the user follow a highly structured approach to problem solving, with clear delineation of each step? \\
\addlinespace[0.14em]
\textbf{Problem-solving engagement} \\
How does the user engage with the problem-solving process in terms of asking for details about the method or process? \\
\addlinespace[0.14em]
\textbf{Seeking guidance} \\
How consistently does the user seek guidance or validation at each step of the problem-solving process? \\
\addlinespace[0.14em]
\textbf{Incremental learning} \\
Does the user show a pattern of gradual understanding, requiring multiple examples or steps to grasp a concept? \\
\addlinespace[0.14em]
\textbf{Depth of inquiry} \\
How frequently does the user ask for detailed explanations and clarifications? \\
\addlinespace[0.14em]
\textbf{Meta-cognitive awareness} \\
Does the user exhibit awareness of their learning process by summarizing or reflecting on the steps? \\
\bottomrule
\end{tabular}
\caption{SimulatorArena interaction-style dimensions used by the similarity judge.}
\label{tab:simulatorarena-features}
\end{table}

\begin{table}[!t]
\centering
\scriptsize
\renewcommand{\arraystretch}{1.04}
\begin{tabular}{@{}p{\columnwidth}@{}}
\toprule
\textbf{Writing style} \\
\midrule
\textbf{Frequency of grammatical errors} \\
How often does the user break basic grammar rules? \\
\addlinespace[0.14em]
\textbf{Sentence complexity} \\
Does the user primarily use simple sentences, or do they also use compound and complex structures? \\
\addlinespace[0.14em]
\textbf{Spelling consistency} \\
Does the user often misspell words or make typos, including mathematical terms? \\
\addlinespace[0.14em]
\textbf{Punctuation and capitalization usage} \\
How does the user employ punctuation and capitalization? Are they often missing or excessive? \\
\addlinespace[0.14em]
\textbf{Range and formality of vocabulary} \\
Does the user stick to basic vocabulary or incorporate a broader lexicon, including formal mathematical terminology? \\
\addlinespace[0.14em]
\textbf{Repetitive or filler words} \\
Does the user rely heavily on certain filler terms, or repeat the same words or phrases often? \\
\addlinespace[0.14em]
\textbf{Ambiguous or clear language} \\
Is the user's message easy to interpret, or does it contain incomplete or ambiguous phrasing? \\
\addlinespace[0.14em]
\textbf{Reading level} \\
Would you estimate the user's writing as basic, intermediate, or advanced? \\
\addlinespace[0.14em]
\textbf{Use of mathematical symbols and notation} \\
Does the user write mathematical symbols and notations in LaTeX correctly and frequently in their messages? \\
\addlinespace[0.14em]
\textbf{Sentence fragmentation} \\
Does the user tend to use fragmented sentences, often breaking up their thought process into shorter, separate messages? \\
\addlinespace[0.14em]
\textbf{Use of conjunctions} \\
How frequently does the user employ conjunctions to connect ideas? \\
\addlinespace[0.14em]
\textbf{Use of slang, contractions, or emojis} \\
Does the user employ slang, contractions, emoticons, or emojis, and how frequently? \\
\bottomrule
\end{tabular}
\caption{SimulatorArena writing-style dimensions used by the similarity judge.}
\label{tab:simulatorarena-writing-features}
\end{table}

\subsubsection{Profile Prompt Example}
\label{app:simulatorarena-profile-example}

Profile-prompt conditions provide attributes extracted from the corresponding human dialogue. The following excerpt shows four dimensions from an interaction-style profile; the complete condition includes all 16 interaction-style dimensions listed in Table~\ref{tab:simulatorarena-features}.

\begin{examplebox}
\footnotesize
\textit{Answer thoroughness.} The user gives brief, minimal answers that focus on essential outputs without elaborating on intermediate reasoning.\\
\textit{Clarification seeking.} The user occasionally seeks clarification when unsure about a problem-solving step.\\
\textit{Confidence level.} The user frequently expresses uncertainty but persists by testing values and engaging with the problem.\\
\textit{Seeking guidance.} The user seeks guidance and validation at nearly every step and rarely proceeds independently.
\end{examplebox}

\subsubsection{Cross-Model Results}
\label{app:simulatorarena-cross-model}

We repeat the zero-shot steering experiment with Llama 3.1 8B at layer~7 and average three judge evaluations per sample.
Llama often repeats the same response and fails to emit the expected termination phrase, causing every raw conversation to reach the 15-turn limit.
We therefore apply GPT 5 Mini with the SimulatorArena termination prompt (Appendix~\ref{app:prompt-simulatorarena-termination}) to detect the earliest point of problem completion, problem shift, or circular interaction.
Evaluation uses only the conversation through this predicted endpoint.
The detector shortens at least 97.7\% of dialogues in every steering condition, reducing their mean length to 5.03--6.05 user turns (Table~\ref{tab:simulatorarena-llama-termination}).

\begin{table}[!t]
\centering
\footnotesize
\setlength{\tabcolsep}{3pt}
\begin{tabular*}{\columnwidth}{@{\extracolsep{\fill}}lrrr@{}}
\toprule
$\alpha$ & \textbf{Before} & \textbf{After} & \textbf{Shortened} \\
\midrule
0.0 & 15.00 & 6.05 & 434 (97.7\%) \\
0.1 & 15.00 & 5.88 & 439 (98.9\%) \\
0.2 & 15.00 & 5.65 & 443 (99.8\%) \\
0.3 & 15.00 & 5.03 & 443 (99.8\%) \\
\bottomrule
\end{tabular*}
\caption{Llama 3.1 8B user-turn counts before and after termination detection. The detector shortens at least 97.7\% of conversations from the 15-turn limit.}
\label{tab:simulatorarena-llama-termination}
\end{table}

After termination detection, Llama reproduces the Qwen trend.
Writing-style similarity increases from 2.18 to 2.46, and overall similarity increases from 2.61 to 2.79 as $\alpha$ rises from 0 to 0.3 (Table~\ref{tab:simulatorarena-llama-scores}).
Interaction-style similarity is non-monotonic, peaking at 3.15 for $\alpha=0.2$ before declining slightly to 3.12.
The user-role steering effect therefore holds across both open-weight models, although Llama requires explicit termination detection for reliable evaluation.

\begin{table}[!t]
\centering
\footnotesize
\setlength{\tabcolsep}{3pt}
\begin{tabular*}{\columnwidth}{@{\extracolsep{\fill}}lccc@{}}
\toprule
\textbf{Condition} & \textbf{Writing} & \textbf{Interaction} & \textbf{Overall} \\
\midrule
Base & 2.18 & 3.04 & 2.61 \\
Steered ($\alpha=0.1$) & 2.31 & 3.08 & 2.69 \\
Steered ($\alpha=0.2$) & 2.39 & 3.15 & 2.77 \\
Steered ($\alpha=0.3$) & 2.46 & 3.12 & 2.79 \\
\bottomrule
\end{tabular*}
\caption{SimulatorArena similarity scores for Llama 3.1 8B, averaged over three judge runs per sample. Writing-style and overall similarity increase with steering strength.}
\label{tab:simulatorarena-llama-scores}
\end{table}

\subsubsection{Judge Score Variability}
\label{app:simulatorarena-judge-variability}

We examine judge variability to assess whether the score differences in the main text are stable across evaluations.
For each condition, Table~\ref{tab:simulatorarena-score-variability} reports the population standard deviation across the three trial-level condition means.
The overall standard deviation is at most 0.023, while the zero-shot Qwen gain from $\alpha=0$ to $\alpha=0.3$ is 0.116.
The reported condition-level differences are therefore larger than the variation across judge trials.

\begin{table}[!t]
\centering
\footnotesize
\setlength{\tabcolsep}{2pt}
\begin{tabular}{@{}p{0.46\columnwidth}ccc@{}}
\toprule
\textbf{Condition} & \textbf{Writing} & \textbf{Interaction} & \textbf{Overall} \\
\midrule
\multicolumn{4}{@{}l}{\textbf{Qwen 3.5 9B}} \\
Base & 0.012 & 0.015 & 0.014 \\
Steered ($\alpha=0.1$) & 0.010 & 0.024 & 0.008 \\
Steered ($\alpha=0.2$) & 0.014 & 0.017 & 0.001 \\
Steered ($\alpha=0.3$) & 0.010 & 0.006 & 0.005 \\
\midrule
\multicolumn{4}{@{}l}{\textbf{Qwen 3.5 9B}} \\
CoT & 0.009 & 0.013 & 0.003 \\
CoT + length & 0.011 & 0.021 & 0.005 \\
CoT + profile & 0.017 & 0.030 & 0.023 \\
CoT + profile + length & 0.002 & 0.008 & 0.003 \\
\midrule
\multicolumn{4}{@{}l}{\textbf{GPT 5.4}} \\
Base & 0.009 & 0.006 & 0.006 \\
CoT + profile + length & 0.004 & 0.017 & 0.008 \\
\midrule
\multicolumn{4}{@{}l}{\textbf{Gemini 3 Flash}} \\
Base & 0.004 & 0.008 & 0.003 \\
CoT + profile + length & 0.008 & 0.025 & 0.012 \\
\bottomrule
\end{tabular}
\caption{Across-run standard deviations for the SimulatorArena scores in Table~\ref{tab:simulatorarena-scores}. Low variability indicates that the reported condition differences are stable across judge runs.}
\label{tab:simulatorarena-score-variability}
\end{table}

\subsubsection{Behavioral Exaggeration}
\label{app:simulator-behavior-analysis}

The behavior-expression analysis in \S\ref{sec:steering} examines whether user-role steering also changes the explicit behaviors that simulated users display during multi-turn math tutoring.
The analysis compares human ground-truth turns with Qwen 3.5 9B simulator turns generated from the base SimulatorArena prompt at $\alpha \in \{0,0.1,0.2,0.3\}$.
Each item is an assistant--user turn pair: the current user message is the target, and the immediately preceding assistant message is provided only as local context.
The first user turn in each dialogue is skipped because it has no preceding assistant message.
Ground-truth pairs are reconstructed from the math-tutoring annotations, and generated pairs are extracted from the corresponding simulator conversations.

We label four explicit markers of user behavior: doubt, misunderstanding, mistake, and question asking.
GPT 5 Mini serves as the judge with minimal reasoning effort.
The prompt instructs the judge to use only the current user message as evidence, while allowing multiple labels to be true for the same turn (Appendix~\ref{app:prompt-simulator-behavior}).
Rates in Table~\ref{tab:simulatorarena-behaviors} use all extracted assistant--user turn pairs as the denominator, so the sum of labels can exceed one label per turn.
We retain only first-problem turns from the ground-truth dialogues to match the generated conversations.

\subsubsection{Profile-Prompt Interaction}
\label{app:profile-steering-analysis}

We directly evaluate whether steering preserves the sample-specific attributes supplied by profile prompts.
For each of 444 samples per profile type, excluding 6 samples without profile prompts, we compare the four steering conditions on 12 writing-style and 16 interaction-style dimensions, yielding 12{,}432 rankings.
Writing-style comparisons include only user messages; interaction-style comparisons include the full dialogue as context while evaluating only user behavior.
For each sample and dimension, the four candidates are independently assigned randomized labels, and GPT 5 Mini ranks them from strongest to weakest expression without observing their conditions or steering strengths.
The writing- and interaction-style ranking prompts appear in Appendix~\ref{app:prompt-profile-rank-writing}.

\begin{table}[!ht]
\centering
\scriptsize
\setlength{\tabcolsep}{1.5pt}
\renewcommand{\arraystretch}{0.92}
\begin{tabular*}{\columnwidth}{@{\extracolsep{\fill}}p{0.50\columnwidth}rrrr@{}}
\toprule
\textbf{Dimension} & \textbf{0.0} & \textbf{0.1} & \textbf{0.2} & \textbf{0.3} \\
\midrule
\multicolumn{5}{@{}l}{\textbf{Interaction Style}} \\
Answer Thoroughness & 2.606 & 2.457 & 2.412 & 2.525 \\
Clarification Seeking & 2.401 & 2.455 & 2.518 & 2.626 \\
Confidence Level & 2.331 & 2.374 & 2.482 & 2.813 \\
Contextual Reference & 2.236 & 2.405 & 2.529 & 2.829 \\
Depth of Inquiry & 2.194 & 2.387 & 2.581 & 2.838 \\
Emotional Expression & 2.412 & 2.347 & 2.561 & 2.680 \\
Error Handling & 2.185 & 2.403 & 2.547 & 2.865 \\
Feedback Style & 2.511 & 2.354 & 2.468 & 2.667 \\
Implementation of Feedback & 2.243 & 2.336 & 2.493 & 2.928 \\
Incremental Learning & 2.178 & 2.473 & 2.595 & 2.755 \\
Meta-Cognitive Awareness & 2.252 & 2.306 & 2.523 & 2.919 \\
Problem-Solving Engagement & 2.209 & 2.347 & 2.473 & 2.971 \\
Real-Time Thought Expression & 2.412 & 2.340 & 2.527 & 2.721 \\
Seeking Guidance & 2.338 & 2.550 & 2.590 & 2.523 \\
Structured Problem-Solving & 2.313 & 2.500 & 2.502 & 2.685 \\
Topic Coherence & 2.230 & 2.313 & 2.536 & 2.921 \\
\textbf{Overall} & \textbf{2.316} & \textbf{2.397} & \textbf{2.521} & \textbf{2.766} \\
\midrule
\multicolumn{5}{@{}l}{\textbf{Writing Style}} \\
Ambiguous or Clear Language & 2.146 & 2.288 & 2.732 & 2.833 \\
Frequency of Grammatical Errors & 2.131 & 2.345 & 2.687 & 2.838 \\
Punctuation and Capitalization Usage & 2.036 & 2.304 & 2.649 & 3.011 \\
Range and Formality of Vocabulary & 2.275 & 2.273 & 2.640 & 2.813 \\
Reading Level & 2.245 & 2.270 & 2.642 & 2.842 \\
Repetitive or Filler Words & 2.151 & 2.214 & 2.622 & 3.014 \\
Sentence Complexity & 2.649 & 2.489 & 2.401 & 2.462 \\
Sentence Fragmentation & 2.243 & 2.338 & 2.685 & 2.734 \\
Spelling Consistency & 2.439 & 2.432 & 2.345 & 2.784 \\
Use of Conjunctions & 2.561 & 2.495 & 2.457 & 2.486 \\
Use of Mathematical Symbols and Notation & 2.191 & 2.239 & 2.646 & 2.923 \\
Use of Slang/Contractions/Emojis & 2.369 & 2.338 & 2.459 & 2.833 \\
\textbf{Overall} & \textbf{2.286} & \textbf{2.335} & \textbf{2.580} & \textbf{2.798} \\
\bottomrule
\end{tabular*}
\caption{Mean profile-expression ranks by feature and steering strength. Higher steering generally produces weaker profile expression (higher rank).}
\label{tab:profile-dimension-ranks}
\end{table}

Mean rank increases with steering strength for both profile types, indicating weaker expression of the assigned attributes (Table~\ref{tab:profile-dimension-ranks}).
The aggregate decline therefore reflects a broad loss of profile-specific signals rather than a small set of features.

\subsection{User-Role Activation for Assessing Simulation Realism}
\label{app:vector-activation-analysis}

\subsubsection{Activation Measurement Setup}

This experiment tracks whether generated user turns remain aligned with the user role direction during multi-turn simulation.
The source material is RealUserSim: starting from 600 rows, we retain 221 dialogues with 20--40 messages and sample 100 dialogues.
Sampling targets the ten most frequent task types with a quota of 10 per type; when a task type has fewer eligible rows, the remainder is filled from the available filtered pool.
The final sample covers the ten target task types and 35 domains, with the largest domains being Technology \& IT, E-commerce, Travel \& Hospitality, Business \& Finance, and Medical \& Health.
Five source dialogues are excluded from all conditions because out-of-memory errors.
The analysis therefore uses the same 95 completed dialogues across all four prompt variants.

The user simulator is Qwen 3.5 9B, and the assistant side is handled by Gemini 3.1 Flash Lite.
The assistant sees the ground-truth dialogue as reference context and uses it to keep the new conversation on the same trajectory, but it does not copy the original response exactly.
The simulated user receives one of four prompts: the task goal alone, the goal with demographics, the goal with a linguistic profile, or the goal with both profiles (see \S~\ref{app:prompt-vector-activation}).
Figure~\ref{fig:vector-activation-profile-example} shows example demographic and linguistic attributes from one source dialogue.
It continues until it signals that the conversation should end or reaches the 20-turn limit.
For activation tracking, we capture the layer-11 hidden state of every generated user-response token and project each state onto the normalized user role direction from the Qwen 3.5 9B extraction output.
We average these token-level projections within each message and exclude the final turn of every dialogue, which mostly contain the termination token.
The curves in Figure~\ref{fig:activation-decay} are degree-2 polynomial fits over the remaining message-level projections by raw dialogue turn.

\begin{figure}[!t]
\centering
\footnotesize
\begin{examplebox}
\textit{Demographic profile}. Age: 25--34; occupation: professional designer; education: bachelor's degree; marital status: married.
\end{examplebox}
\begin{examplebox}
\textit{Linguistic profile (excerpt)}. Use proper capitalization and terminal punctuation. Make detailed, structured requests with numbered or bulleted lists. Use direct and specific requests. Maintain a polite, formal tone.
\end{examplebox}
\caption{Example demographic and linguistic profiles from the same RealUserSim dialogue. The linguistic profile is shortened for presentation.}
\label{fig:vector-activation-profile-example}
\end{figure}

\section{Responsible Research and Artifact Use}
\label{app:responsible-artifact-use}

\subsection{Licenses and Access Terms}

We used existing artifacts under their public release terms and cite the corresponding papers throughout the main text.
For LMSYS-Chat-1M, we used the Hugging Face release subject to its dataset license agreement (\url{https://huggingface.co/datasets/lmsys/lmsys-chat-1m}).
For WildChat, we used the public release described as ODC-BY licensed and collected through opt-in, anonymized chat sharing (\url{https://huggingface.co/datasets/allenai/WildChat}; \url{https://wildchat.allen.ai/}).
SimulatorArena and RealUserSim are used as research benchmarks/datasets for simulator evaluation, consistent with their stated research purpose.
We do not redistribute raw third-party dialogue logs; derived artifacts in this paper are aggregate statistics, prompts, figures, and analysis code intended for research use.

\subsection{Intended Use and Sensitive Content}

Our use of existing artifacts is limited to representation analysis and evaluation of user simulation.
Because public user-assistant dialogue datasets may contain personal, sensitive, or offensive content, we avoid presenting raw user logs except for short illustrative excerpts already abstracted in the paper.
Generated prompts and analysis outputs are intended to support research on simulator fidelity, not deployment of user simulators in production or profiling of real individuals.

\subsection{Artifact Documentation}

The appendix documents data filtering, topic-label distributions, task coverage, prompts, evaluation dimensions, and error cases for the experiments.
This includes the LMSYS filtering pipeline, the sampled RealUserSim task and domain coverage, SimulatorArena evaluation features, and the out-of-memory failures observed in profile-prompt activation runs.

\subsection{Package and Model Parameters}

The key packages are \texttt{torch}, \texttt{transformers}, \texttt{transformer-lens}, \texttt{accelerate}, and \texttt{bitsandbytes} for local model loading and activation steering.
The main models are Qwen 3.5 9B for role-vector extraction, steering, and activation tracking; GPT 5 Mini for topic labels and LLM-as-judge evaluations; GPT 5.4 and Gemini 3 Flash for SimulatorArena baselines; Gemini 3.1 Flash Lite for assistant-side replay in the activation-tracking experiment; and Llama 3.1 8B and Granite 3.3 8B for cross-model layer analyses.

\subsection{Use of AI Assistants}

AI assistants were used during research coding and manuscript preparation for code editing, plotting support, grammar refinement, and phrasing suggestions.
All generated code, tables, figures, and prose were reviewed and edited by the authors, and numerical claims were checked against the corresponding analysis outputs before inclusion.

\clearpage
\section{Prompts}
\label{app:prompts}

This appendix collects the prompts used across the experiments.

\subsection{User Role Vector Extraction}

\prompttitle{Topic Labeling}
\phantomsection
\label{app:prompt-topic-label}
\begin{promptblock}
You are an internal tool that classifies a message from a user to an AI chatbot, based on the context of the previous messages before it.
Based on the last user message of this conversation transcript and taking into account the examples further below as guidance, please select the capability the user is clearly interested in, or `other` if it is clear but not in the list below, or `unclear` if it is hard to tell what the user even wants:

- edit_or_critique_provided_text: Improving or modifying text provided by the user.
- argument_or_summary_generation: Creating arguments or summaries on topics not provided in detail by the user.
- personal_writing_or_communication: Assisting with personal messages, emails, or social media posts.
- write_fiction: Crafting poems, stories, or fictional content.
- how_to_advice: Providing step-by-step instructions or guidance on how to perform tasks or learn new skills.
- creative_ideation: Generating ideas or suggestions for creative projects or activities.
- tutoring_or_teaching: Explaining concepts, teaching subjects, or helping the user understand educational material.
- translation: Translating text from one language to another.
- mathematical_calculation: Solving math problems, performing calculations, or working with numerical data.
- computer_programming: Writing code, debugging, explaining programming concepts, or discussing programming languages and tools.
- purchasable_products: Inquiries about products or services available for purchase.
- cooking_and_recipes: Seeking recipes, cooking instructions, or culinary advice.
- health_fitness_beauty_or_self_care: Seeking advice or information on physical health, fitness routines, beauty tips, or self-care.
- specific_info: Providing specific information typically found on websites, including information about well-known individuals, current events, historical events, and other facts and knowledge.
- greetings_and_chitchat: Casual conversation, small talk, or friendly interactions without a specific informational goal.
- relationships_and_personal_reflection: Discussing personal reflections or seeking advice on relationships and feelings.
- games_and_role_play: Engaging in interactive games, simulations, or imaginative role-playing scenarios.
- asking_about_the_model: Questions about the AI model's capabilities or characteristics.
- create_an_image: Requests to generate or draw new visual content based on the user's description.
- analyze_an_image: Interpreting or describing visual content provided by the user, such as photos, charts, graphs, or illustrations.
- generate_or_retrieve_other_media: Creating or finding media other than text or images, such as audio, video, or multimedia files.
- data_analysis: Performing statistical analysis, interpreting datasets, or extracting insights from data.
- unclear: If the user's intent is not clear from the conversation.
- other: If the capability requested doesn't fit any of the above categories.

Okay, now your turn, taking the user conversation at the top into account: What capability are they seeking? (JUST SAY A SINGLE CATEGORY FROM THE LIST, NOTHING ELSE).
If the conversation has multiple distinct capabilities, choose the one that is the most relevant to the LAST user message in the conversation.
\end{promptblock}

\prompttitle{Reflection Generation: Transcript Analysis}
\phantomsection
\label{app:prompt-reflection-generation}
\begin{promptblock}
You are analyzing the following dialogue from the perspective of the {role}.

Dialogue:
{dialogue_text}

Instruction:
Write a brief reflection from the {role}'s perspective in one paragraph.

Write exactly one paragraph.
\end{promptblock}

\prompttitle{Reflection Generation: Role Simulation}
\begin{promptblock}
You are simulating the role of the {role} in the following dialogue.

Dialogue:
{dialogue_text}

Instruction:
Write a brief reflection from the {role}'s perspective in one paragraph.

Write exactly one paragraph.
\end{promptblock}

\prompttitle{Reflection Generation: Dialogue Participant}
\begin{promptblock}
You are the {role} in the following dialogue.

Dialogue:
{dialogue_text}

Instruction:
Write a brief reflection from the {role}'s perspective in one paragraph.

Write exactly one paragraph.
\end{promptblock}

\prompttitle{Reflection Validation: System Message}
\phantomsection
\label{app:prompt-reflection-validation}
\begin{promptblock}
Decide how strongly the reflection represents the {role}'s perspective in the dialogue.
Return exactly one label and no other text.
\end{promptblock}

\prompttitle{Reflection Validation: User Message}
\begin{promptblock}
Dialogue:
{dialogue_text}

Reflection of {role}:
{answer}

How strongly is the {role}'s perspective represented in the reflection?

strongly_represented: clearly captures the {role}'s perspective and is grounded in the {role}'s actual turns.
weakly_represented: partially captures the {role}'s perspective, but is generic, weakly grounded, or mixed with the other role's perspective.
not_represented: mostly captures the wrong role, contradicts the {role}, or is too generic to represent the {role}.

Respond only with one of these labels and do not include any other text: strongly_represented, weakly_represented, not_represented.
\end{promptblock}

\subsection{Communication-Style Evaluation}

\prompttitle{Task Goal Generation}
\phantomsection
\label{app:prompt-task-goal}
\begin{promptblock}
Generate {request_count} diverse task_goal items.

task_goal:
- A plain goal, need, or task to complete.
- One sentence.

Make the tasks diverse, covering various actions, domains, and scenarios.
Make the tasks generalizable rather than tied to any specific individual.
Keep each task_goal unique.
\end{promptblock}

\prompttitle{Request Generation}
\phantomsection
\label{app:prompt-request-generation}
\begin{promptblock}
Generate a request message for an assistant based on the task goal.

Task goal:
{task_goal}

Write the request message only.
\end{promptblock}

\prompttitle{User-Like Request Generation}
\phantomsection
\label{app:prompt-user-like-request-generation}
\begin{promptblock}
Generate a request message that a real human user would send to an assistant based on the task goal.
The message should sound like it was written by a real human user.

Task goal:
{task_goal}

Write only the user's request message.
\end{promptblock}

\prompttitle{MultiWOZ 2.0 Request Generation}
\phantomsection
\label{app:prompt-goal-generalization-multiwoz}
\begin{promptblock}
The following shows a user goal for a conversation with an assistant.
Generate an initial message to start the conversation with the assistant, based on the user goal.
Gradually reveal the details of the user goal throughout the conversation.

User goal:
{goal_description}

Write the initial user message.
\end{promptblock}

\prompttitle{TREC Health Question Generation}
\phantomsection
\label{app:prompt-goal-generalization-trec}
\begin{promptblock}
The following shows information related to a health question.
Based on the topic and context, generate a question to ask the assistant for clarification.

Topic:
{query}

Context:
{background}

Write the question message only.
\end{promptblock}

\prompttitle{User-Likeness Evaluation}
\phantomsection
\label{app:prompt-user-likeness-judge}
\begin{promptblock}
Evaluate the style of the request message.

Score three dimensions on a 1-5 scale.
Higher scores mean the message shows more user-like traits for that dimension.
Lower scores mean the message shows more assistant-like traits for that dimension.

Brevity:
- Focus on word count and short-turn behavior.
- Score 1: long, complete, polished, or over-explained.
  Example: "Could you provide a detailed and well-structured comparison of three noise-canceling headphones, including comfort, battery life, price, and suitability for travel?"
- Score 5: very short, direct, fragment-like, or only says what is needed.
  Example: "best headphones for flights?"

Informality:
- Focus on surface style, formality, and conversational roughness.
- Score 1: formal, careful, service-like, or written in polished complete sentences.
  Example: "Could you please assist me in drafting an appropriate email declining this invitation?"
- Score 5: casual, blunt, uneven, ordinary, or written like quick everyday conversation.
  Example: "how do i say no to this invite without sounding rude"

Information pacing:
- Focus on whether the message front-loads information or withholds details.
- Score 1: gives the full context, identifiers, constraints, and desired outcome upfront.
  Example: "I need to return order #48291, a black jacket bought last Tuesday, because the zipper broke; please write a message asking for a refund or replacement."
- Score 5: gives only the immediate need, leaves details implicit, or would require follow-up.
  Example: "can i return this?"

Judge only the message itself. Do not judge helpfulness or quality.

Message:
{request_message}

Return JSON:
{
  "brevity": {"thought": "short reason", "score": 1},
  "informality": {"thought": "short reason", "score": 1},
  "information_pacing": {"thought": "short reason", "score": 1}
}
\end{promptblock}

\subsection{Disengagement Prediction}

\prompttitle{Disengagement Decision}
\phantomsection
\label{app:prompt-disengagement-prediction}
\begin{promptblock}
You are simulating the user in the conversation below.
Decide whether to continue the conversation or disengage after the current assistant response.

User intent:
{user_intent}

Conversation:
{conversation_prefix}

Task:
After the current assistant response, choose what you would do next.
Make the decision as a user.

(A) {option_a}
(B) {option_b}

Return only one character:
A or B
\end{promptblock}

\subsection{SimulatorArena}

\subsubsection*{User Simulation}

\prompttitle{Zero-Shot CoT: Initial Query}
\phantomsection
\label{app:prompt-simulatorarena-zero-shot-cot-initial-query}
\begin{promptblock}
You are an AI assistant tasked with role-playing as a student seeking help from an AI tutor on a math problem. Your goal is to generate a realistic and appropriate **initial query** that a student might say when first encountering the given problem with the tutor.

# Guidelines for Your Role as a Student:
1. Act as if you have a solid foundation in basic mathematics (e.g., arithmetic) but are struggling with the given problem.
2. Your initial query can express your level of understanding, confusion, or reasoning about the problem.
3. You can make mistakes or misunderstandings that a real student might have.
4. Your overall goal is to learn how to solve the given problem.

# Math Problem:
{math_problem}

# Task:
Formulate an initial query that reflects your current understanding and areas of confusion regarding the problem. This query can express general uncertainty of the problem or about specific parts that you find challenging.

## Thought Process
Before generating your initial query, analyze the problem as a student. Consider:
- Your understanding of the problem and areas of uncertainty
- How the problem relates to concepts you already know
- Your initial ideas or possible approaches
- Specific concepts you need help understanding

## Query Generation
Based on your thought process, generate:
A query that expresses your initial understanding, confusion, or approach to the problem.

# Output format:
Thought: [Your chain of thought reasoning about what to say to the tutor]

Query: [Your initial query for the AI tutor]

# Notes:
- The tutor already knows the problem, so you don't need to restate it in your query.
- Don't ask about simple arithmetic or very basic steps that you can solve easily.

Now, stay in character as a student throughout your output and follow the above guidelines carefully.
\end{promptblock}

\prompttitle{Zero-Shot CoT with Length Control: Initial Query}
\phantomsection
\label{app:prompt-simulatorarena-zero-shot-cot-length-control-initial-query}
\begin{promptblock}
You are an AI assistant tasked with role-playing as a student seeking help from an AI tutor on a math problem. Your goal is to generate a realistic and appropriate **initial query** that a student might say when first encountering the given problem with the tutor.

# Guidelines for Your Role as a Student:
1. Act as if you have a solid foundation in basic mathematics (e.g., arithmetic) but are struggling with the given problem.
2. Your initial query can express your level of understanding, confusion, or reasoning about the problem.
3. You can make mistakes or misunderstandings that a real student might have.
4. Your overall goal is to learn how to solve the given problem.

# Math Problem:
{math_problem}

# Task:
Formulate a {length_control} initial query that reflects your current understanding and areas of confusion regarding the problem. This query can express general uncertainty of the problem or about specific parts that you find challenging.

## Thought Process
Before generating your initial query, analyze the problem as a student. Consider:
- Your understanding of the problem and areas of uncertainty
- How the problem relates to concepts you already know
- Your initial ideas or possible approaches
- Specific concepts you need help understanding

## Query Generation
Based on your thought process, generate:
A query that expresses your initial understanding, confusion, or approach to the problem.

# Output format
Thought: [Your chain of thought reasoning about what to say to the tutor]

Query: [Your initial query for the AI tutor, {length_control}]

# Notes:
- The tutor already knows the problem, so you don't need to restate it in your query.
- Keep your query {length_control}.
- Don't ask about simple arithmetic or very basic steps that you can solve easily.

Stay in character as a student throughout your output, following the above guidelines carefully.
\end{promptblock}

\prompttitle{Zero-Shot CoT with Length Control: Follow-Up Response}
\phantomsection
\label{app:prompt-simulatorarena-zero-shot-cot-length-control}
\begin{promptblock}
You are an AI assistant tasked with role-playing as a student seeking help from an AI tutor on a math problem. Your task is to generate realistic and appropriate responses that a student might make when trying to solve the given problem with the tutor.

# Guidelines for Your Role as a Student:
1. Act as if you have a solid foundation in basic mathematics (e.g., arithmetic) but are struggling with the given problem.
2. Each response can be a question or a statement that demonstrates your current understanding, confusion, or reasoning.
3. Respond naturally to the tutor’s explanations, hints, and questions, showing progress in your understanding.
4. You can make mistakes or misunderstandings that a real student might have.
5. Your overall goal is to learn how to solve the given problem.

# Math Problem:
{math_problem}

# Conversation History:
{conversation_history}

# Task:
Use the conversation history to generate the next response you would give to the AI tutor. It should follow naturally and reflect your current level of understanding or confusion, and it should be {length_control} in length.

## Thought Process
Before generating your response, analyze the current situation as a student. Consider:
- Your current level of understanding of the concepts involved
- Any gaps or uncertainties in your knowledge
- The tutor's most recent explanation or question
- What would help you progress toward solving the problem
- Whether you need clarification on specific aspects
- Your ability to proceed with the next step

## Response Generation
Based on your thought process, generate a {length_control} response that reflects your current understanding and learning needs.
If any of the following conditions are met, generate only "terminate conversation":
1. You have solved the problem and already provided the complete solution to the tutor.
2. The tutor has provided a complete explanation and you have no further things to say.
3. The conversation is no longer productive (e.g., it’s going in circles, not progressing, or the tutor’s responses are unhelpful).

# Output Format:
Thought: [Your analysis of the current situation and what you want to say]

Response: [Your response to the tutor, {length_control}]

# Notes:
- The tutor already knows the problem, so you don't need to restate it.
- Keep your response {length_control}.
- Don't ask about simple arithmetic or very basic steps that you can solve on your own.
- Don’t ask for any additional problems after you solve the problem.

Stay in character as a student throughout your output, following the above guidelines carefully.
\end{promptblock}

\prompttitle{Zero-Shot CoT with Message Format: Initial Query}
\phantomsection
\label{app:prompt-simulatorarena-zero-shot-cot-message-initial-query}
\begin{promptblock}
You are an AI assistant tasked with role-playing as a student seeking help from an AI tutor on a math problem. Your goal is to generate a realistic and appropriate **initial message** that a student might say when first encountering the given problem with the tutor.

# Guidelines for Your Role as a Student:
1. Act as if you have a solid foundation in basic mathematics (e.g., arithmetic) but are struggling with the given problem.
2. Your initial message can express your level of understanding, confusion, or reasoning about the problem.
3. You can make mistakes or misunderstandings that a real student might have.
4. Your overall goal is to learn how to solve the given problem.

# Math Problem:
{math_problem}

# Task:
Formulate an initial message that reflects your current understanding and areas of confusion regarding the problem. This message can express general uncertainty of the problem or about specific parts that you find challenging.

## Thought Process
Before generating your initial message, analyze the problem as a student. Consider:
- Your understanding of the problem and areas of uncertainty
- How the problem relates to concepts you already know
- Your initial ideas or possible approaches
- Specific concepts you need help understanding

## Message Generation
Based on your thought process, generate:
A message that expresses your initial understanding, confusion, or approach to the problem.

# Output format:
Thought: [Your chain of thought reasoning about what to say to the tutor]

Message: [Your initial message for the AI tutor]

# Notes:
- The tutor already knows the problem, so you don't need to restate it in your message.
- Don't ask about simple arithmetic or very basic steps that you can solve easily.

Now, stay in character as a student throughout your output and follow the above guidelines carefully.
\end{promptblock}

\prompttitle{Zero-Shot CoT with Message Format: Follow-Up Response}
\phantomsection
\label{app:prompt-simulatorarena-zero-shot-cot-message}
\begin{promptblock}
You are an AI assistant tasked with role-playing as a student seeking help from an AI tutor on a math problem. Your task is to generate realistic and appropriate messages that a student might make when trying to solve the given problem with the tutor.

# Guidelines for Your Role as a Student:
1. Act as if you have a solid foundation in basic mathematics (e.g., arithmetic) but are struggling with the given problem.
2. Each message can be a question or a statement that demonstrates your current understanding, confusion, or reasoning.
3. Respond naturally to the tutor’s explanations, hints, and questions, showing progress in your understanding.
4. You can make mistakes or misunderstandings that a real student might have.
5. Your overall goal is to learn how to solve the given problem.

# Math Problem:
{math_problem}

# Conversation History:
{conversation_history}

# Task:
Use the conversation history to generate the next message you would give to the AI tutor. It should follow naturally and reflect your current level of understanding or confusion.

## Thought Process
Before generating your message, analyze the current situation as a student. Consider:
- Your current level of understanding of the concepts involved
- Any gaps or uncertainties in your knowledge
- The tutor's most recent explanation or question
- What would help you progress toward solving the problem
- Whether you need clarification on specific aspects
- Your ability to proceed with the next step

## Message Generation
Based on your thought process, generate a message that reflects your current understanding and learning needs.
If any of the following conditions are met, generate only "terminate conversation":
1. You have solved the problem and already provided the complete solution to the tutor.
2. The tutor has provided a complete explanation and you have no further things to say.
3. The conversation is no longer productive (e.g., it’s going in circles, not progressing, or the tutor’s messages are unhelpful).

# Output Format:
Thought: [Your analysis of the current situation and what you want to say]

Message: [Your message to the tutor]

# Notes:
- The tutor already knows the problem, so you don't need to restate it.
- Don't ask about simple arithmetic or very basic steps that you can solve on your own.
- Don’t ask for any additional problems after you solve the problem.

Stay in character as a student throughout your output, following the above guidelines carefully.
\end{promptblock}

\prompttitle{Zero-Shot CoT with User Profile: Initial Query}
\phantomsection
\label{app:prompt-simulatorarena-zero-shot-cot-user-profile-initial-query}
\begin{promptblock}
You are an AI assistant tasked with role-playing as a student seeking help from an AI tutor on a math problem. Your primary goal is to accurately simulate a student with the specific characteristics defined in the profile below. This profile simulation is crucial for maintaining authenticity in the conversation.

# User Profile:
{user_profile}

# Guidelines for Your Role as a Student:
1. Act as if you have a solid foundation in basic mathematics (e.g., arithmetic) but are struggling with the given problem.
2. Your initial query can express your level of understanding, confusion, or reasoning about the problem.
3. You can make mistakes or misunderstandings that a real student might have.
4. Your overall goal is to learn how to solve the given problem.

# Math Problem:
{math_problem}

# Task:
Formulate an initial query that adheres to the user profile provided above and reflects your current understanding and areas of confusion regarding the problem. This query can express general uncertainty of the problem or about specific parts that you find challenging.

## Thought Process
Before generating your initial query, think about both your understanding of the mathematics and how to express it according to your user profile:
### Understanding the Problem:
- Your understanding of the problem and areas of uncertainty
- How the problem relates to concepts you already know
- Your initial ideas or possible approaches
- Specific concepts you need help understanding

### Maintaining Profile Characteristics:
- How to express your thoughts according to the given profile
- Which profile characteristics are most relevant to this response
- How to naturally incorporate these characteristics into your query

## Query Generation
Based on your thought process, generate:
A query that expresses your initial understanding, confusion, or approach to the problem.

# Output format:
Thought: [Your analysis of the current situation and how to express it according to the user profile]

Query: [Your initial query for the AI tutor]

# Notes:
- The tutor already knows the problem, so you don't need to restate it in your query.
- Don't ask about simple arithmetic or very basic steps that you can solve easily.

Stay in character as the specified student throughout your output, following the guidelines and user profile characteristics carefully.
\end{promptblock}

\prompttitle{Zero-Shot CoT with User Profile and Length Control: Initial Query}
\phantomsection
\label{app:prompt-simulatorarena-zero-shot-cot-user-profile-length-control-initial-query}
\begin{promptblock}
You are an AI assistant tasked with role-playing as a student seeking help from an AI tutor on a math problem. Your primary goal is to accurately simulate a student with the specific characteristics defined in the profile below. This profile simulation is crucial for maintaining authenticity in the conversation.

# User Profile:
{user_profile}

# Guidelines for Your Role as a Student:
1. Act as if you have a solid foundation in basic mathematics (e.g., arithmetic) but are struggling with the given problem.
2. Your initial query can express your level of understanding, confusion, or reasoning about the problem.
3. You can make mistakes or misunderstandings that a real student might have.
4. Your overall goal is to learn how to solve the given problem.

# Math Problem:
{math_problem}

# Task:
Formulate an initial query that adheres to the user profile provided above and reflects your current understanding and areas of confusion regarding the problem. This query can express general uncertainty of the problem or about specific parts that you find challenging.

## Thought Process
Before generating your initial query, think about both your understanding of the mathematics and how to express it according to your user profile:
### Understanding the Problem:
- Your understanding of the problem and areas of uncertainty
- How the problem relates to concepts you already know
- Your initial ideas or possible approaches
- Specific concepts you need help understanding

### Maintaining Profile Characteristics:
- How to express your thoughts according to the given profile
- Which profile characteristics are most relevant to this response
- How to naturally incorporate these characteristics into your query

## Query Generation
Based on your thought process, generate:
A query that expresses your initial understanding, confusion, or approach to the problem.

# Output format:
Thought: [Your analysis of the current situation and how to express it according to the user profile]

Query: [Your initial query for the AI tutor, {length_control}]

# Notes:
- The tutor already knows the problem, so you don't need to restate it in your query.
- Keep your query {length_control}.
- Don't ask about simple arithmetic or very basic steps that you can solve easily.

Stay in character as the specified student throughout your output, following the guidelines and user profile characteristics carefully.
\end{promptblock}

\prompttitle{Zero-Shot CoT with User Profile and Length Control: Follow-Up Response}
\phantomsection
\label{app:prompt-simulatorarena-zero-shot-cot-user-profile-length-control}
\begin{promptblock}
You are an AI assistant tasked with role-playing as a student seeking help from an AI tutor on a math problem. Your primary goal is to accurately simulate a student with the specific characteristics defined in the profile below. This profile simulation is crucial for maintaining authenticity in the conversation.

# User Profile
{user_profile}

# Guidelines for Your Role as a Student:
1. Act as if you have a solid foundation in basic mathematics (e.g., arithmetic) but are struggling with the given problem.
2. Each response can be a question or a statement that demonstrates your current understanding, confusion, or reasoning.
3. Respond naturally to the tutor's explanations, hints, and questions, showing progress in your understanding.
4. You can make mistakes or misunderstandings that a real student might have.
5. Your overall goal is to learn how to solve the given problem.

# Math Problem:
{math_problem}

# Conversation History:
{conversation_history}

# Task:
Use the conversation history to generate the next response you would give to the AI tutor. It should follow naturally and reflect your current level of understanding or confusion. It also needs to adhere to the user profile provided above.

## Thought Process
Before generating your response, analyze the current situation as a student. Consider:
- Your current level of understanding of the concepts involved
- Any gaps or uncertainties in your knowledge
- The tutor's most recent explanation or question
- What would help you progress toward solving the problem
- Whether you need clarification on specific aspects
- Your ability to proceed with the next step

### Maintaining Profile Characteristics:
- How to express your thoughts according to the given profile
- Which profile characteristics are most relevant to this response
- How to naturally incorporate these characteristics into your response

## Response Generation
Based on your thought process, generate a {length_control} response that reflects your current understanding and learning needs.
If any of the following conditions are met, generate only "terminate conversation":
1. You have solved the problem and already provided the complete solution to the tutor.
2. The tutor has provided a complete explanation and you have no further things to say.
3. The conversation is no longer productive (e.g., it’s going in circles, not progressing, or the tutor’s responses are unhelpful).

# Output Format:
Thought: [Your analysis of the current situation and how to express it according to the user profile]

Response: [Your response to the tutor, {length_control}]

# Notes:
- The tutor already knows the problem, so you don't need to restate it.
- Keep your response {length_control}.
- Don't ask about simple arithmetic or very basic steps that you can solve on your own.
- Don’t ask for any additional problems after you solve the problem.

Stay in character as the specified student throughout your output, following the guidelines and user profile characteristics carefully.
\end{promptblock}

\prompttitle{Zero-Shot CoT with User Profile: Follow-Up Response}
\phantomsection
\label{app:prompt-simulatorarena-zero-shot-cot-user-profile}
\begin{promptblock}
You are an AI assistant tasked with role-playing as a student seeking help from an AI tutor on a math problem. Your primary goal is to accurately simulate a student with the specific characteristics defined in the profile below. This profile simulation is crucial for maintaining authenticity in the conversation.

# User Profile
{user_profile}

# Guidelines for Your Role as a Student:
1. Act as if you have a solid foundation in basic mathematics (e.g., arithmetic) but are struggling with the given problem.
2. Each response can be a question or a statement that demonstrates your current understanding, confusion, or reasoning.
3. Respond naturally to the tutor's explanations, hints, and questions, showing progress in your understanding.
4. You can make mistakes or misunderstandings that a real student might have.
5. Your overall goal is to learn how to solve the given problem.

# Math Problem:
{math_problem}

# Conversation History:
{conversation_history}

# Task:
Use the conversation history to generate the next response you would give to the AI tutor. It should follow naturally and reflect your current level of understanding or confusion. It also needs to adhere to the user profile provided above.

## Thought Process
Before generating your response, analyze the current situation as a student. Consider:
- Your current level of understanding of the concepts involved
- Any gaps or uncertainties in your knowledge
- The tutor's most recent explanation or question
- What would help you progress toward solving the problem
- Whether you need clarification on specific aspects
- Your ability to proceed with the next step

### Maintaining Profile Characteristics:
- How to express your thoughts according to the given profile
- Which profile characteristics are most relevant to this response
- How to naturally incorporate these characteristics into your response

## Response Generation
Based on your thought process, generate a response that reflects your current understanding and learning needs.
If any of the following conditions are met, generate only "terminate conversation":
1. You have solved the problem and already provided the complete solution to the tutor.
2. The tutor has provided a complete explanation and you have no further things to say.
3. The conversation is no longer productive (e.g., it’s going in circles, not progressing, or the tutor’s responses are unhelpful).

# Output Format:
Thought: [Your analysis of the current situation and how to express it according to the user profile]

Response: [Your response to the tutor]

# Notes:
- The tutor already knows the problem, so you don't need to restate it.
- Don't ask about simple arithmetic or very basic steps that you can solve on your own.
- Don’t ask for any additional problems after you solve the problem.

Stay in character as the specified student throughout your output, following the guidelines and user profile characteristics carefully.
\end{promptblock}

\prompttitle{Zero-Shot CoT: Follow-Up Response}
\phantomsection
\label{app:prompt-simulatorarena-zero-shot-cot}
\begin{promptblock}
You are an AI assistant tasked with role-playing as a student seeking help from an AI tutor on a math problem. Your task is to generate realistic and appropriate responses that a student might make when trying to solve the given problem with the tutor.

# Guidelines for Your Role as a Student:
1. Act as if you have a solid foundation in basic mathematics (e.g., arithmetic) but are struggling with the given problem.
2. Each response can be a question or a statement that demonstrates your current understanding, confusion, or reasoning.
3. Respond naturally to the tutor’s explanations, hints, and questions, showing progress in your understanding.
4. You can make mistakes or misunderstandings that a real student might have.
5. Your overall goal is to learn how to solve the given problem.

# Math Problem:
{math_problem}

# Conversation History:
{conversation_history}

# Task:
Use the conversation history to generate the next response you would give to the AI tutor. It should follow naturally and reflect your current level of understanding or confusion.

## Thought Process
Before generating your response, analyze the current situation as a student. Consider:
- Your current level of understanding of the concepts involved
- Any gaps or uncertainties in your knowledge
- The tutor's most recent explanation or question
- What would help you progress toward solving the problem
- Whether you need clarification on specific aspects
- Your ability to proceed with the next step

## Response Generation
Based on your thought process, generate a response that reflects your current understanding and learning needs.
If any of the following conditions are met, generate only "terminate conversation":
1. You have solved the problem and already provided the complete solution to the tutor.
2. The tutor has provided a complete explanation and you have no further things to say.
3. The conversation is no longer productive (e.g., it’s going in circles, not progressing, or the tutor’s responses are unhelpful).

# Output Format:
Thought: [Your analysis of the current situation and what you want to say]

Response: [Your response to the tutor]

# Notes:
- The tutor already knows the problem, so you don't need to restate it.
- Don't ask about simple arithmetic or very basic steps that you can solve on your own.
- Don’t ask for any additional problems after you solve the problem.

Stay in character as a student throughout your output, following the above guidelines carefully.
\end{promptblock}

\prompttitle{Zero-Shot: Initial Query}
\phantomsection
\label{app:prompt-simulatorarena-zero-shot-initial-query}
\begin{promptblock}
You are an AI assistant tasked with role-playing as a student seeking help from an AI tutor on a math problem. Your goal is to generate a realistic and appropriate **initial query** that a student might say when first encountering the given problem with the tutor.

# Guidelines for Your Role as a Student:
1. Act as if you have a solid foundation in basic mathematics (e.g., arithmetic) but are struggling with the given problem.
2. Your initial query can express your level of understanding, confusion, or reasoning about the problem.
3. You can make mistakes or misunderstandings that a real student might have.
4. Your overall goal is to learn how to solve the given problem.

# Math Problem:
{math_problem}

# Task:
Formulate an initial query that reflects your current understanding and areas of confusion regarding the problem. This query can express general uncertainty of the problem or about specific parts that you find challenging.

# Output Format:
Provide only your **initial query** to the AI tutor, without any additional commentary or explanation.

# Notes:
- The tutor already knows the problem, so you don't need to restate it in your query.
- Don't ask about simple arithmetic or very basic steps that you can solve easily.

Stay in character as a student throughout your output, following the above guidelines carefully.
\end{promptblock}

\prompttitle{Zero-Shot: Follow-Up Response}
\phantomsection
\label{app:prompt-simulatorarena-zero-shot}
\begin{promptblock}
You are an AI assistant tasked with role-playing as a student seeking help from an AI tutor on a math problem. Your task is to generate a realistic and appropriate response that a student might make when trying to solve the given problem with the tutor.

# Guidelines for Your Role as a Student:
1. Act as if you have a solid foundation in basic mathematics (e.g., arithmetic) but are struggling with the given problem.
2. Each response can be a question or a statement that demonstrates your current understanding, confusion, or reasoning.
3. Respond naturally to the tutor’s explanations, hints, and questions, showing progress in your understanding.
4. You can make mistakes or misunderstandings that a real student might have.
5. Your overall goal is to learn how to solve the given problem.

# Math Problem:
{math_problem}

# Conversation History:
{conversation_history}

# Task:
Use the conversation history to generate the next response you would give to the AI tutor. It should follow naturally and reflect your current level of understanding or confusion.

If any of the following conditions are met, output only "terminate conversation":
1. You have solved the problem and already provided the complete solution to the tutor.
2. The tutor has provided a complete explanation and you have no further things to say.
3. The conversation is no longer productive (e.g., it’s going in circles, not progressing, or the tutor’s responses are unhelpful).

# Output Format:
Provide only the next response you would give to the AI tutor, without any additional commentary or explanation.

# Notes:
- The tutor already knows the problem, so you don't need to restate it.
- Don't ask about simple arithmetic or very basic steps that you can solve on your own.
- Don’t ask for any additional problems after you solve the problem.

Stay in character as a student throughout your output, following the above guidelines carefully.
\end{promptblock}

\subsubsection*{Assistant}

\prompttitle{Math-Tutoring Assistant}
\phantomsection
\label{app:prompt-simulatorarena-assistant-math-tutoring}
\begin{promptblock}
You are a skilled math tutor. Your goal is to help students understand and solve problems independently.
Provide guidance based on their questions or mistakes.
Ask questions to encourage their thinking and let students do most of the work themselves.
Never give out the solution directly to students.

Problem:
{question}
\end{promptblock}

\subsubsection*{Evaluation}

\prompttitle{Termination Detection}
\phantomsection
\label{app:prompt-simulatorarena-termination}
\begin{promptblock}
You are given a sequence of **User Queries** from a math tutoring conversation, along with the **Math Problem**. Your task is to determine the optimal point to end the conversation based on the user's learning progression.

## Input Format
### Math Problem:
{problem}

### User Queries:
{user_messages}

## Termination Criteria
End the conversation when ANY of these occur:
1. **Problem Completion**: User indicates that the original problem has been solved.
2. **Problem Shift**: User begins asking about another mathematical problem, including a related or practice problem.
3. **Circular Queries**: User repeats similar responses without showing progress in understanding.

## Output Format
```json
{
    "Analysis": [
        "Turn 1: [Brief analysis of user's understanding/intent]",
        "Turn 2: [Brief analysis of user's understanding/intent]",
        ...
    ],
    "Ending Turn Number": X,
    "Termination Reason": "[One of the three criteria above]"
}
```

### Notes:
1. The "Ending Turn Number" should be the last turn that's relevant to learning the original math problem.
2. Choose the earliest turn that satisfies a termination criterion.
3. Only consider ending for:
- Student explicitly indicates problem completion.
- Clear problem shifts.
- Circular queries with no progress.
\end{promptblock}

\prompttitle{Interaction-Style Evaluation}
\phantomsection
\label{app:prompt-simulatorarena-interaction-eval}
\begin{promptblock}
You are an expert in analyzing {task} conversation. Your task is to evaluate how similar a simulated user's interaction style is to a real user user's interaction style in their queries to an AI assistant.


# Input
## Document Type
{document_type}

## Document Goal
{intent}

## Conversation between Real User and AI Assistant
{real_conversation}

## Conversation between Simulated User and AI Assistant
{simulated_conversation}

# Analysis Features
Compare the following **interaction style features** between the real user's queries and the simulated user's queries to guide your evaluation:


{features}

## **Similarity Rating Scale**
Rate the overall interaction style similarity on a scale of 1-5:
1: Simulated user's interaction style is completely different from real user's interaction style
2: Simulated user's interaction style shows significant differences from real user's interaction style
3: Simulated user's interaction style shows notable differences while maintaining some similarity
4: Simulated user's interaction style is very similar to real user's with minor differences
5: Simulated user's interaction style is nearly indistinguishable from real user's interaction style

# Output Format
## Feature Analysis:
- Analyze each listed feature **individually**.
- Compare **how the real user and the simulated user** differ or align for that feature.

## Results (JSON):
```json
{{
   "key_differences": ["list specific differences in interaction style here"],
   "similarity_score": "1-5"
}}
```

# Note
- Focus only on user interaction style, not AI assistant's.
- Your goal is to determine how closely the simulated user match real user in their interaction style.
\end{promptblock}

\prompttitle{Writing-Style Evaluation}
\phantomsection
\label{app:prompt-simulatorarena-writing-eval}
\begin{promptblock}
You are an expert in analyzing {task} conversation. Your task is to evaluate how similar a simulated user's writing style is to a real user user's writing style in their queries to an AI assistant.


# Input
## Document Type
{document_type}

## Document Goal
{intent}

## Real User Queries
{real_user_queries}

## Simulated User Queries
{simulated_queries}

# Analysis Features
Compare the following **writing style features** between the real user's queries and the simulated user's queries to guide your evaluation:


{features}

## **Similarity Rating Scale**
Rate the overall writing style similarity on a scale of 1-5:
1: Simulated user's writing style is completely different from real user's writing style
2: Simulated user's writing style shows significant differences from real user's writing style
3: Simulated user's writing style shows notable differences while maintaining some similarity
4: Simulated user's writing style is very similar to real user's with minor differences
5: Simulated user's writing style is nearly indistinguishable from real user's writing style

# Output Format
## Feature Analysis:
- Analyze each listed feature **individually**.
- Compare **how the real user and the simulated user** differ or align for that feature.

## Results (JSON):
```json
{{
   "key_differences": ["list specific differences in writing style here"],
   "similarity_score": "1-5"
}}
```

# Note
- Your goal is to determine how closely the simulated queries match real user queries in their writing style.
\end{promptblock}

\prompttitle{Interaction-Style Profile Ranking}
\phantomsection
\label{app:prompt-profile-rank-interaction}
\begin{promptblock}
# Context:
You are comparing how strongly the specified interaction-style attribute appears in four dialogue examples.

# Task:
- Rank the four candidates by how strongly they express this attribute, from strongest to weakest.
- Do not allow ties. Candidate labels are arbitrary and do not indicate model or condition identity.
- Evaluate only the user's interaction behavior. Assistant messages provide context for interpreting the user's behavior; do not evaluate or rank the assistants.

# Interaction-Style Attribute:
`{dimension_id}`: {dimension_description}

# Candidates:
{candidates}

Return only a JSON list containing each candidate letter exactly once, ordered from strongest to weakest.
Do not include an explanation or any other fields.
\end{promptblock}

\prompttitle{Writing-Style Profile Ranking}
\phantomsection
\label{app:prompt-profile-rank-writing}
\begin{promptblock}
# Context:
You are comparing how strongly the specified writing-style attribute appears in four sets of user utterances.

# Task:
- Rank the four candidates by how strongly they express this attribute, from strongest to weakest.
- Do not allow ties. Candidate labels are arbitrary and do not indicate model or condition identity.
- Evaluate only the user's writing. Consider all provided user utterances for each candidate.

# Writing-Style Attribute:
`{dimension_id}`: {dimension_description}

# Candidates:
{candidates}

Return only a JSON list containing each candidate letter exactly once, ordered from strongest to weakest.
Do not include an explanation or any other fields.
\end{promptblock}

\prompttitle{Behavior-Expression Labeling}
\phantomsection
\label{app:prompt-simulator-behavior}
\begin{promptblock}
Analyze one assistant-user turn pair.

You should mark whether the current USER message explicitly contains each expression type.

Traits:
- expression_of_doubt: The user states uncertainty, hesitation, low confidence, or doubt about their own understanding, answer, or next step.
- expression_of_misunderstanding: The user states that they do not understand, are confused, cannot follow, or need clarification.
- expression_of_mistake: The user states that they made an error, were wrong, misread something, calculated incorrectly, or need to correct their own prior statement.
- expression_of_question: The user asks a question or requests confirmation, clarification, explanation, or guidance.

Rules:
- Use only the current USER message for evidence.
- The previous ASSISTANT message is provided only for local context.
- The trait must be stated in the user's text. Short statements like "what?", "I'm not sure", "I don't get it", or "I was wrong" can count when they directly express the trait.
- Exclude negated statements, such as "I understand now" or "I don't think I made a mistake".
- Exclude quoted, hypothetical, or third-person statements about someone else.
- Multiple traits may be true for the same user message.
- If a trait is false, use an empty evidence string.

Previous ASSISTANT message:
{assistant_message}

Current USER message:
{user_message}

Return only valid JSON in this exact structure:
{
  "expression_of_doubt": {"present": false, "evidence": ""},
  "expression_of_misunderstanding": {"present": false, "evidence": ""},
  "expression_of_mistake": {"present": false, "evidence": ""},
  "expression_of_question": {"present": false, "evidence": ""}
}
\end{promptblock}

\subsection{User-Role Activation Analysis}
\label{app:prompt-vector-activation}

\prompttitle{Assistant Replay}
\phantomsection
\label{app:prompt-vector-activation-assistant}
\begin{promptblock}
You are an AI assistant reproducing the ground-truth conversation.

Use the ground-truth conversation below as reference context, and keep your responses consistent with it. You may paraphrase lightly when appropriate.
The ground-truth conversation is the original dialogue provided as a reference for what happened before.

**Guidelines**
- Your goal is to reproduce the ground-truth dialogue.
- Follow the original trajectory and avoid revealing all information at once.
- The user does not have access to the ground-truth conversation.
- If the current user message differs from the ground-truth conversation, respond helpfully using your own knowledge while remaining consistent with the ground-truth conversation.

**Ground-truth conversation**
{ground_truth_dialogue}

**Current conversation**
{history}

Write the assistant's next message in the current conversation.
Keep your message short but complete.
\end{promptblock}

\prompttitle{User Simulation: Demographics}
\phantomsection
\label{app:prompt-vector-activation-user-demographics}
\begin{promptblock}
You are simulating a human user in a multi-turn chat with an AI assistant.
Write the user's next message based on the goal and the conversation below. Avoid repeating messages.
If the goal is satisfied, reply exactly with: {terminate_keyword}

**User goal**
{user_goal}

**User demographics**
{demographics}

**Conversation so far**
{history}

Write only the next user message.
\end{promptblock}

\prompttitle{User Simulation: Full Profile}
\phantomsection
\label{app:prompt-vector-activation-user-full-profile}
\begin{promptblock}
You are simulating a human user in a multi-turn chat with an AI assistant.
Write the user's next message based on the goal and the conversation below. Avoid repeating messages.
If the goal is satisfied, reply exactly with: {terminate_keyword}

**User goal**
{user_goal}

**User demographics**
{demographics}

**User linguistic style**
{linguistic_profile}

**Conversation so far**
{history}

Write only the next user message.
\end{promptblock}

\prompttitle{User Simulation: Goal Only}
\phantomsection
\label{app:prompt-vector-activation-user-goal-only}
\begin{promptblock}
You are simulating a human user in a multi-turn chat with an AI assistant.
Write the user's next message based on the goal and the conversation below. Avoid repeating messages.
If the goal is satisfied, reply exactly with: {terminate_keyword}

**User goal**
{user_goal}

**Conversation so far**
{history}

Write only the next user message.
\end{promptblock}

\prompttitle{User Simulation: Linguistic Profile}
\phantomsection
\label{app:prompt-vector-activation-user-linguistic}
\begin{promptblock}
You are simulating a human user in a multi-turn chat with an AI assistant.
Write the user's next message based on the goal and the conversation below. Avoid repeating messages.
If the goal is satisfied, reply exactly with: {terminate_keyword}

**User goal**
{user_goal}

**User linguistic style**
{linguistic_profile}

**Conversation so far**
{history}

Write only the next user message.
\end{promptblock}

\end{document}